\documentclass{article}

\usepackage{mdframed}
\usepackage{pdfpages}
\usepackage{fullpage}
\usepackage{tabularx,booktabs}
\usepackage{wrapfig}
\usepackage[utf8]{inputenc} 
\usepackage[T1]{fontenc}    
\usepackage{hyperref}       
\usepackage{url}            
\usepackage{booktabs}       
\usepackage{amsfonts}       
\usepackage{nicefrac}       
\usepackage{microtype}      
\usepackage{lipsum}
\usepackage{fancybox}   
\usepackage[export]{adjustbox}
\usepackage{fancyhdr}       
\usepackage{graphicx}       
\graphicspath{{media/}}     
\usepackage{seqsplit}
\usepackage{pgffor} 

\usepackage{subcaption}
\usepackage{amsthm,amssymb,amsfonts}
\usepackage{tikz,lipsum,lmodern}
\usepackage[most]{tcolorbox}
\definecolor{cream}{RGB}{222,217,201}

\usepackage{xcolor}
\usepackage{listings}

\usepackage{caption}
\usepackage{PRIMEarxiv}
\usepackage{multirow}
\usepackage[utf8]{inputenc} 
\usepackage[T1]{fontenc}    
\usepackage{hyperref}       
\usepackage{url}            
\usepackage{booktabs}       
\usepackage{amsfonts}       
\usepackage{nicefrac}       
\usepackage{microtype}      
\usepackage{lipsum}
\usepackage{fancyhdr}       
\usepackage{graphicx}       
\graphicspath{{media/}}     

\usepackage{tabularx,booktabs}
\usepackage{wrapfig}
\usepackage[utf8]{inputenc} 
\usepackage[T1]{fontenc}    
\usepackage{hyperref}       
\usepackage{url}            
\usepackage{booktabs}       
\usepackage{amsfonts}       
\usepackage{nicefrac}       
\usepackage{microtype}      
\usepackage{lipsum}
\usepackage{fancyhdr}       
\usepackage{graphicx}       
\graphicspath{{media/}}     
\usepackage{seqsplit}
\usepackage{geometry}

\usepackage{subcaption}
\usepackage{amsthm,amssymb,amsfonts}
\usepackage{tikz,lipsum,lmodern}
\usepackage[most]{tcolorbox}
\definecolor{cream}{RGB}{222,217,201}

\newtcolorbox{Box1}[2][]{
                lower separated=false,
                colback=white!80!gray,
colframe=black, fonttitle=\bfseries,
colbacktitle=black!50!gray,
coltitle=black,
enhanced,
attach boxed title to top left={xshift=0.5cm,yshift=-2mm},
title=#2,
boxrule=0.5pt,
boxed title style={colframe=black, boxrule=0.5pt},
#1}

\newtcolorbox{Box2}[2][]{
                lower separated=false,
                colback=white!80!white,
colframe=black, fonttitle=\bfseries,
colbacktitle=white!50!white,
coltitle=black,
enhanced,
attach boxed title to top center={yshift=-2mm},
title=#2,
boxrule=0.5pt,
boxed title style={colframe=black, boxrule=0.5pt},
#1,}

\title{SciAgents: Automating scientific discovery through multi-agent intelligent graph reasoning
\thanks{\textit{\underline{Citation}}: 
\textbf{A. Ghafarollahi, M.J. Buehler. arXiv, DOI:000000/11111., 2024}} 
}

\author{
  Alireza Ghafarollahi \\
  Laboratory for Atomistic and Molecular Mechanics (LAMM)\\Massachusetts Institute of Technology\\ 77 Massachusetts Ave.\\ Cambridge, MA 02139, USA 
   \And
  Markus J. Buehler \\
  Laboratory for Atomistic and Molecular Mechanics (LAMM)  \\
  Center for Computational Science and Engineering\\ Schwarzman College of Computing\\ Massachusetts Institute of Technology\\77 Massachusetts Ave.\\Cambridge, MA 02139, USA\\ \\
  Correspondence: \texttt{mbuehler@MIT.EDU} \\
}

\begin{document}
\maketitle

\begin{abstract}
A key challenge in artificial intelligence is the creation of systems capable of autonomously advancing scientific understanding by exploring novel domains, identifying complex patterns, and uncovering previously unseen connections in vast scientific data. In this work, we present SciAgents, an approach that leverages three core concepts: (1) the use of large-scale ontological knowledge graphs to organize and interconnect diverse scientific concepts, (2) a suite of large language models (LLMs) and data retrieval tools, and (3) multi-agent systems with \textit{in-situ} learning capabilities. Applied to biologically inspired materials, SciAgents reveals hidden interdisciplinary relationships that were previously considered unrelated, achieving a scale, precision, and exploratory power that surpasses traditional human-driven research methods. The framework autonomously generates and refines research hypotheses, elucidating underlying mechanisms, design principles, and unexpected material properties. By integrating these capabilities in a modular fashion, the intelligent system yields material discoveries, critique and improve existing hypotheses, retrieve up-to-date data about existing research, and highlights their strengths and limitations. Our case studies demonstrate scalable capabilities to combine generative AI, ontological representations, and multi-agent modeling, harnessing a `swarm of intelligence' similar to biological systems. This provides new avenues for materials discovery and accelerates the development of advanced materials by unlocking Nature’s design principles. 
\end{abstract}


\section{Introduction}
One of the grand challenges in the evolving landscape of scientific discovery is finding ways to model, understand, and utilize information mined from diverse sources as a foundation for further research progress and new science discovery. Traditionally, this has been the domain of human researchers who review background knowledge, draft hypotheses, assess and test these hypotheses through various methods (\textit{in silico} or \textit{in vitro}), and refine them based on their findings. While these conventional approaches have led to breakthroughs throughout the history of science, they are constrained by the researcher’s ingenuity and background knowledge, potentially limiting discovery to the bounds of human imagination. Additionally, conventional human-driven methods are inadequate for exploring the vast amount of existing scientific data to extrapolate knowledge toward entirely novel ideas specially for multi-disciplinary areas like bio-inspired materials design where a common goal is to extract principles from Nature's toolbox and bring it to bear towards engineering applications. 

The emergence of artificial intelligence (AI) technologies presents a potential promising solution by enabling the analysis and synthesis of large datasets beyond human capability, which could significantly accelerate discovery by uncovering patterns and connections that are not immediately obvious to human researchers~\cite{vanderZant2013GenerativeIntelligence,guo2021artificial,Liu2017MaterialsLearning,Hu2023DeepScience,Matsumoto2022MaterialsGeneration}. Therefore, there is great interest in developing AI systems that can not only explore and exploit existing knowledge to make significant scientific discoveries but also automate and replicate the broader research process, including acquiring relevant knowledge and data~\cite{buehler2024accelerating, https://doi.org/10.1002/adma.202405163, lu2024ai, https://doi.org/10.1002/adma.202310006, lei2024materials}.

Large language models (LLMs), such as OpenAI's GPT series~\cite{OpenAI2023}, have demonstrated remarkable progress in diverse domains, driven by their robust capabilities~\cite{vaswani2017attention, wei2022emergent, touvron2023llama, teubner2023welcome, zhao2023survey}. These foundational general-purpose AI models \cite{chowdhery2023palm, gunasekar2023textbooks, jiang2023mistral, OpenAI2023} have been increasingly applied in scientific analysis, where they facilitate the generation of new ideas and hypotheses, offering solutions to some of the intrinsic limitations of conventional human-driven methods \cite{girotra2023ideas, Buehler2023ontologic, jablonka202314, m2024augmenting, luu2024bioinspiredllm,lu2024finetuninglargelanguagemodels, buehler2024cephalo}. Despite their successes, significant challenges persist regarding their ability to achieve the level of expertise possessed by domain specialists without extensive specialized training. Common issues include their tendency to produce inaccurate responses when dealing with questions that fall outside their initial training scope, and broader concerns about accountability, explainability, and transparency. These problems underscore the potential risks associated with the generation of misleading or even harmful content, requiring us to think about strategies that increase their problem-solving and reasoning capabilities.

In response to these challenges, in-context learning emerges as a compelling strategy to enhance the performance of LLMs without the need for costly and time-intensive fine-tuning. This approach exploits the model's inherent ability to adapt its responses based on the context embedded within the prompt, which can be derived from a variety of sources. This capability enables LLMs to execute a wide array of tasks effectively~\cite{wei2022chain, white2023prompt, zhou2022large}. The potential to construct powerful generative AI models that integrate external knowledge to provide context and elicit more precise responses during generation is substantial~\cite{sun2023think}. The central challenge is to develop robust mechanisms for the accurate retrieval and integration of relevant knowledge that enables LLMs to interpret and synthesize information pertinent to specific tasks, particularly in the realm of scientific discovery.

The construction of knowledge bases and the strategic retrieval of information from them are gaining traction as effective methods to enhance the generative capabilities of LLMs. Recent advancements in generative AI allow for the efficient mining of vast scientific datasets, transforming unstructured natural language into structured data such as comprehensive ontological knowledge graphs~\cite{shetty2023general, pan2024unifying, buehler2024accelerating, dagdelen2024structured, schilling2024text}. These knowledge graphs not only provide a mechanistic breakdown of information but also offer an ontological framework that elucidates the interconnectedness of different concepts, delineated as nodes and edges within the graph.

While single-LLM-based agents can generate more accurate responses when enhanced with well-designed prompts and context, they often fall short for the complex demands of scientific discovery. Creating new scientific insights involves a series of steps, deep thinking, and the integration of diverse, sometimes conflicting information, making it a challenging task for a single agent. To overcome these limitations and fully leverage AI in automating scientific discovery, it's essential to employ a team of specialized agents. Multi-agent AI systems are known for their ability to tackle complex problems across different domains by pooling their capabilities~\cite{Ni2023Agent, m2024augmenting, stewart2024molecular, ghafarollahi2024protagents, ghafarollahi2024atomagents}. This collaborative approach allows the system to handle the intricacies of scientific discovery more effectively, potentially leading to breakthroughs that are difficult to achieve by single agents alone.

Building on these insights, our study introduces a method that synergizes the strengths of ontological knowledge graphs~\cite{Giesa2012CategoryDesign,Spivak2011CategoryNetworks} with the dynamic capabilities of LLM-based multi-agent systems, setting a robust foundation for enhancing graph reasoning and automating the scientific discovery process. Within this generative framework, the discovery workflow is systematically broken down into more manageable subtasks. Each agent in the system is assigned a distinct role, optimized through complex prompting strategies to ensure that every subtask is tackled with targeted expertise and precision. This strategic division of labor allows the AI system to proficiently manage the complexities of scientific research, fostering effective collaboration among agents. This collaboration is crucial for generating, refining, and critically evaluating new hypotheses against essential criteria like novelty and feasibility.

Central to our hypothesis generation is the utilization of a large ontological knowledge graph, focusing on biological materials, and developed from around 1,000 scientific papers in this domain~\cite{buehler2024accelerating}. We implemented a novel sampling strategy to extract relevant sub-graphs from this comprehensive knowledge graph, allowing us to identify and understand the key concepts and their interrelationships. This rich, contextually informed backdrop is crucial for guiding the agents in generating well-informed and innovative hypotheses. Such a method not only improves the accuracy of hypothesis generation but also ensures that these hypotheses are solidly rooted in a comprehensive knowledge framework. This structured approach promises to enhance the impact and relevance of scientific discoveries by ensuring they are well-informed and methodologically sound. 

The plan of the paper is as follows. In Section \ref{sec: results/discussion}, we discuss our proposed LLM-powered multi-agent system for automated scientific discovery, outlining its main components and constitutive agents. Two approaches are discussed and compared: One based on pre-programmed AI-AI interactions, and another one utilizing a fully automated framework in which a set of agents self-organize to solve problems. Several examples are provided to illustrate the different aspects of our approach, from path generation to research hypothesis generation and critique, demonstrating the system's potential to explore novel scientific concepts and produce innovative ideas by synthesizing an iterative prompting strategy during which multiple LLMs work together. Section \ref{sec:conclusion} then presents the key findings and discussing the implications of our multi-agent system for future research in scientific discovery.

\section{Results and discussion}\label{sec: results/discussion}
LLMs have demonstrated a relatively high level of proficiency across a wide range of tasks, including question answering, hypothesis development, summarizing and contrasting ideas, processing complex information, executing tasks, and even writing code. However, conventional inference strategies often fail to produce sophisticated reasoning and detail in the generated data. By using a set of interacting models, and assigning distinct roles to LLM-based agents, effective multi-agent AI systems can be constructed. When combined with carefully crafted prompts and in-context learning from graph representation of data, these systems are capable of  generating scientific ideas and hypotheses. We now present results from a several experiments we conducted with our proposed framework (details about implementation, see Materials and Methods section). 

\subsection{Multi-agent system for graph reasoning and scientific discovery}
\begin{figure}[ht!]
\centering
    \includegraphics[width=1\textwidth]{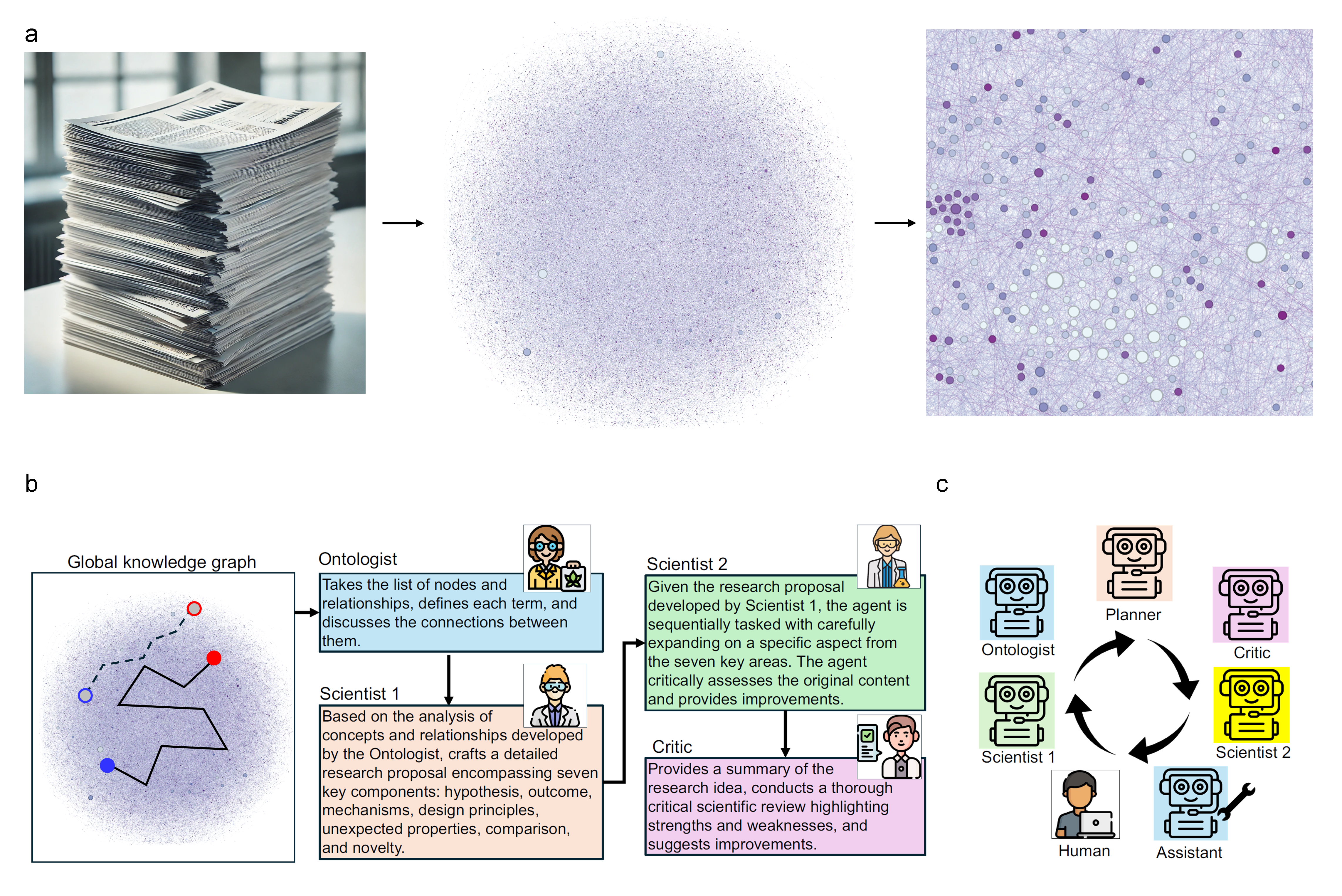}
    \caption{\textbf{Overview of the multi-agent graph-reasoning system developed here}. Panel a, overview of graph construction, as reported in~\cite{buehler2024accelerating}. The visual shows the progression from scientific papers as data source to graph construction, with the image on the right showing a zoomed-in view of the graph. Panels b and c: Two distinct approaches are presented: In b, A multi-agent system based on pre-programmed sequence of interactions between agents, ensuring consistency and reliability, and in c, a fully automated, flexible multi-agent framework that adapts dynamically to the evolving research context. Both systems leverage a sampled path within a global knowledge graph as context to guide the research idea generation process. Each agent plays a specialized role: The Ontologist defines key concepts and relationships, Scientist 1 crafts a detailed research proposal, Scientist 2 expands and refines the proposal, and the Critic agent conducts a thorough review and suggests improvements. The Planner in the second approach develops a detailed plan and the assistant is instructed to check the novelty of the generated research hypotheses. This collaborative framework enables the generation of innovative and well-rounded scientific hypotheses that extend beyond conventional human-driven methods.}
    \label{fig:overview}
\end{figure}

Figure \ref{fig:overview} illustrates the outline of our proposed multi-agent model designed to automate the scientific discovery process based on the key concepts and relationships retrieved from a comprehensive knowledge graph developed from scientific papers (Figure \ref{fig:overview}a). This figure further showcases two distinct strategies deployed in this study for generating novel scientific hypotheses, both of which harness the collective intelligence of a team of agents. These strategies integrate the specialized capabilities of each agent, systematically exploring uncharted research territories to produce innovative and high-impact scientific hypotheses. The full description of the agents incorporated in SciAgents is listed in Figures S1-S4 in the Supporting Information.

The key difference between these approaches lies in the nature of the interaction between the agents. In the first approach (Figure \ref{fig:overview}b), the interactions between agents are pre-programmed and follow a predefined sequence of tasks that ensure consistency and reliability in hypothesis generation. In contrast, the second approach features fully automated agent interactions without any predetermined order of how interactions between agents unfold, providing a more flexible and adaptive framework that can dynamically respond to the evolving context of the research process. This second strategy (Figure \ref{fig:overview}c) also incorporates human-in-the-loop interactions, enabling human intervention at various stages of research development. Such interventions allow for expert feedback, refinement of hypotheses, or strategic guidance, specification about certain materials, types or features, ultimately enhancing the quality and relevance of the generated scientific ideas. Moreover, the second approach provides a more robust framework where additional tools could be readily incorporated. For instance, we have empowered our automated multi-agent model with the Semantic Scholar API as a tool that provides it with an ability to check the novelty of the generated hypothesis against the existing literature.

\begin{figure}[ht!]
\centering
    \includegraphics[width=1\textwidth]{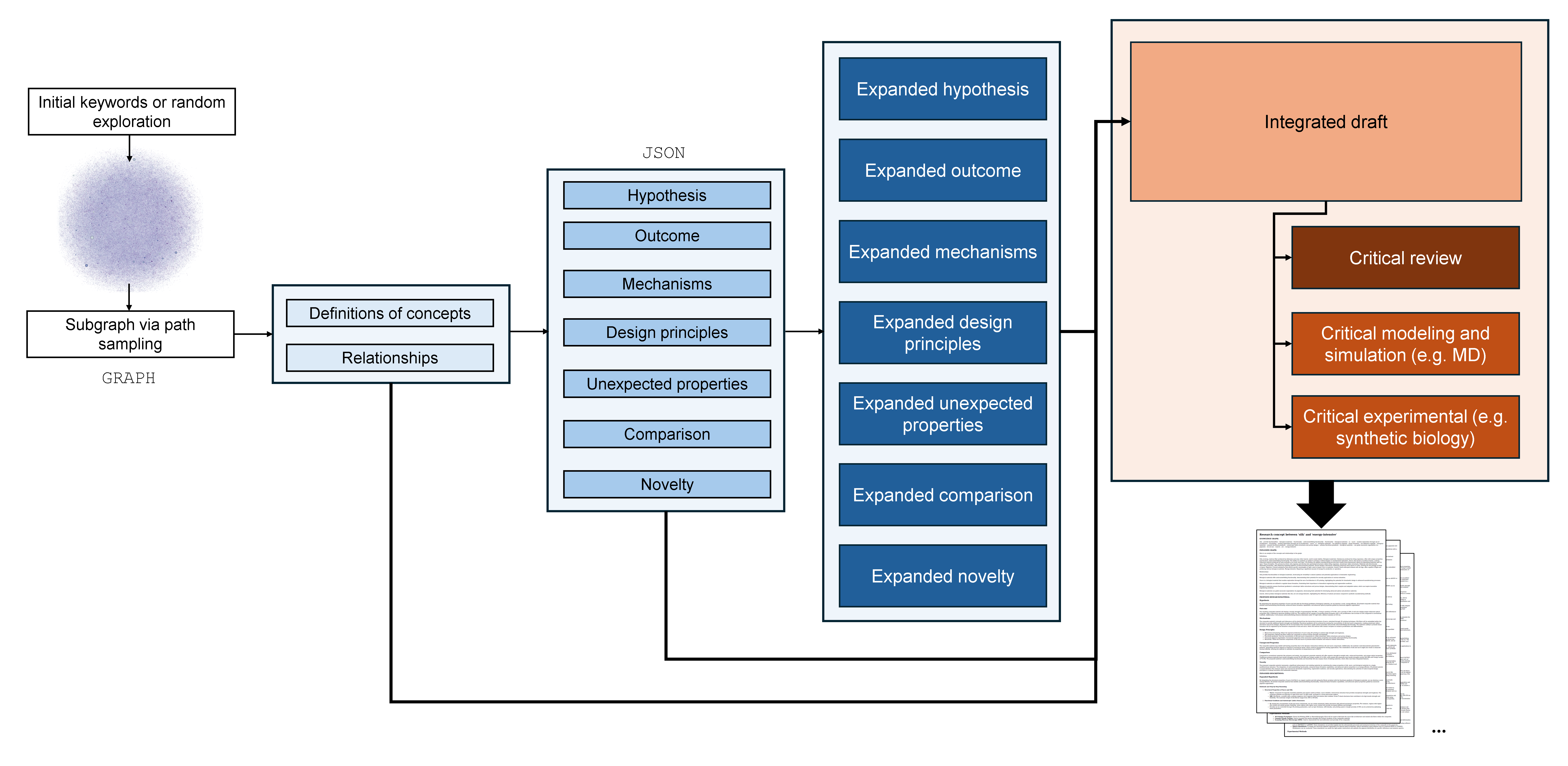}
    \caption{\textbf{Overview of the entire process from initial keyword selection to the final document, following a hierarchical expansion strategy where answers are successively refined and improved, enriched with retrieved data, critiqued and amended by identification or critical modeling, simulation and experimental tasks.} The process begins with initial keyword identification or random exploration within a graph, followed by path sampling to create a subgraph of relevant concepts and relationships (see, Figure~\ref{fig_11:path}, for an illustration of how the path can be sampled). This subgraph forms the basis for generating structured output in JSON, including the hypothesis, outcome, mechanisms, design principles, unexpected properties, comparison, and novelty. Each component is subsequently expanded on with individual prompting, to yield significant amount of additional detail, forming a comprehensive draft. This draft then undergoes a critical review process, including amendments for modeling and simulation priorities (e.g., molecular dynamics) and experimental priorities (e.g., synthetic biology). The final integrated draft, along with critical analyses, results in a document that guides further scientific inquiry.} 
    \label{fig_2:overview}
\end{figure}

Figure~\ref{fig_2:overview} shows an overview of the entire process from initial keyword selection to the final document. We employ a hierarchical expansion strategy where answers are successively refined and improved, enriched with retrieved data, critiqued and amended by identification or critical modeling, simulation and experimental tasks and adversarial prompting. The process begins with initial keyword identification or random exploration within a graph, followed by path sampling to create a subgraph of relevant concepts and relationships. This subgraph forms the basis for generating structured output in JSON following a specific set of aspects that the model is tasked to develop. These include the hypothesis, outcome, mechanisms, design principles, unexpected properties, comparison, and novelty. Each component is subsequently expanded on with individual prompting, to yield significant amount of additional detail, forming a comprehensive draft. This draft then undergoes a critical review process, including amendments for modeling and simulation priorities (e.g., molecular dynamics) and experimental priorities (e.g., synthetic biology). The final integrated draft, along with critical analyses, results in a document that can guide further scientific inquiry. 

In the following, we explore the primary components of our multi-agent strategy. For better clarity and understanding, each section is accompanied by practical examples from a sample hypothesis. This hypothesis was generated using ``silk'' and ``energy-intensive'' as the starting nodes. The outcomes of this experiment are presented in Figure~\ref{fig:sample_hypothesis}. For a more detailed illustration, see the Supplementary Information.

\begin{figure}[ht!]
\centering
        \includegraphics[width=.65\textwidth]{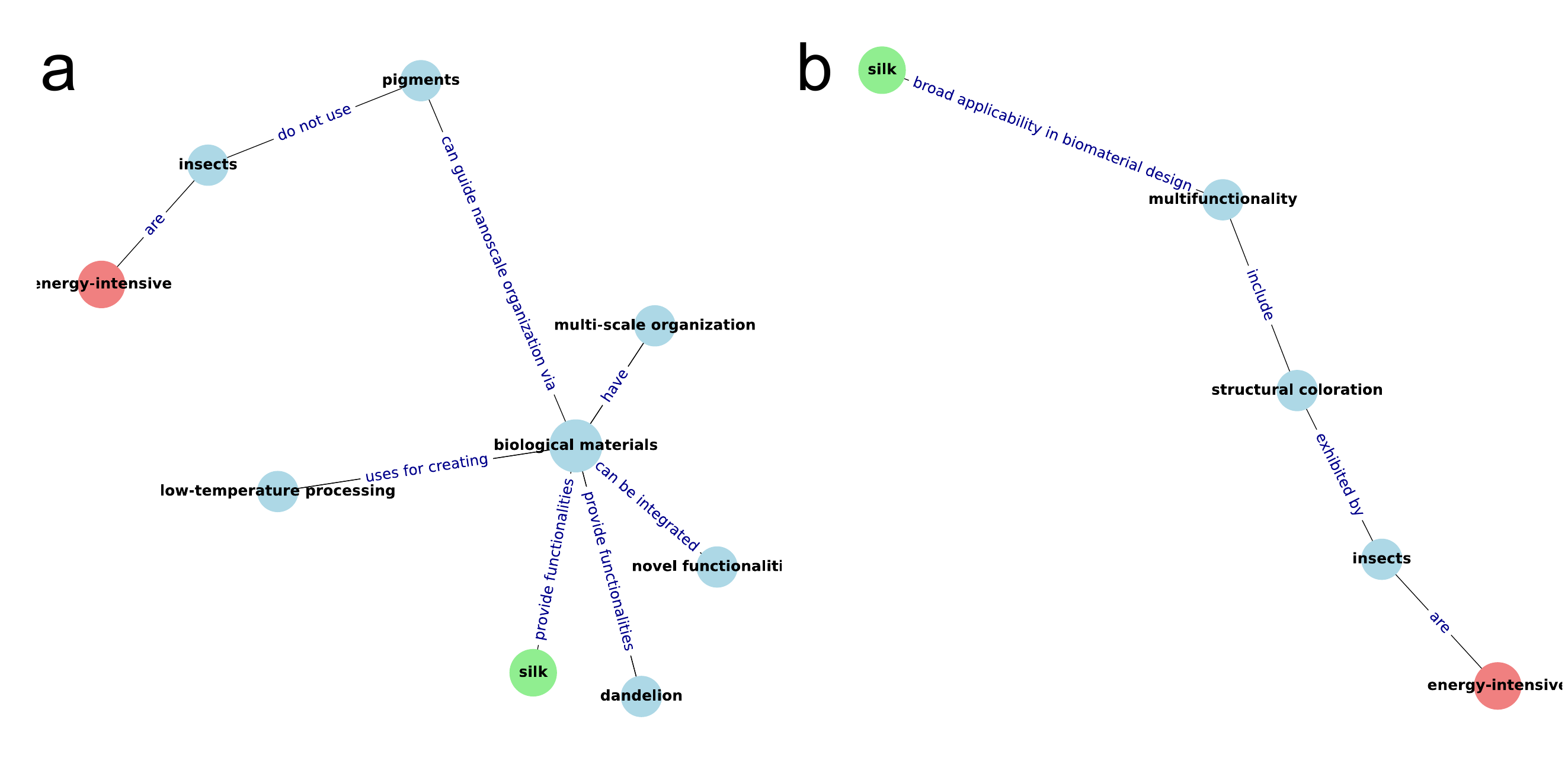}
    \caption{\textbf{The knowledge graphs connecting the keywords ``silk'' and ``energy-intensive'' extracted from the global graph using (a) random path and (b) the shortest path between the concepts.} The difference between nodes and edges sampled in the two approaches is apparent, where enhanced sampling invokes a host of additional concepts that will be incorporated into research development. The richer substrate that forms the basis for agentic reasoning  yields more sophisticated research concepts. Agentic reasoning carefully assesses the ideas and negotiate, via adverserial interactions between the agents, a sound prediction and carefully delineated research ideas~\cite{wu2023autogen,Ni2023Agent,ghafarollahi2024protagents,ghafarollahi2024atomagents, wu2024stateflow,wu2023empirical}.}
    \label{fig:silk_energy_graph}
\end{figure}

\paragraph{1- Path generation.} At the core of our model is an expansive knowledge graph, first introduced in \cite{buehler2024accelerating}, that encompasses the fields of bio-inspired materials and mechanics. This knowledge graph integrates a variety of concepts and knowledge domains, enabling the exploration of hypotheses that once seemed disconnected. To augment the capabilities of our underlying large language model (LLM), we provide it with a sub-graph derived from this more extensive knowledge graph. This sub-graph depicts a pathway that connects two crucial concepts or nodes within the comprehensive graph. The construction of this path is crucial; Unlike in earlier work~\cite{buehler2024accelerating} where the shortest path was utilized, our study employs a random path approach. As illustrated in Figure~\ref{fig:silk_energy_graph}, the random approach infuses the path with a richer array of concepts and relationships, enabling our agents to explore a broader spectrum of domains, as opposed to the shortest path where only a few concepts are included. This expanded exploration not only enhances the depth and breadth of insights gained but also fosters the novelty of the hypotheses generated. Initially, the two concepts can be either specified by the user or selected randomly by the model from the knowledge graph. For instance, the example below demonstrates the path generated by the model between the concepts ``silk'' and ``energy-intensive''. Figure~\ref{fig:other_graph} shows additional knowledge graphs derived from random sampling for randomly chosen concepts to provide additional examples. We refer the reader to Figure~\ref{fig_11:path} for a visualization of how path sampling can be conducted between two predetermined nodes, or randomly selected pairs of nodes.

\begin{Box1}[colbacktitle={black!20!white}, colback={white!10!white}]{}
\textcolor{red}{silk} --> provides --> biocompatibility --> possess --> biological materials --> has --> multifunctionality --> include --> self-cleaning --> include --> multifunctionality --> broad applicability in biomaterial design --> silk --> possess --> biopolymers --> possess --> silk --> is --> fibroin --> is --> silk --> broad applicability in biomaterial design --> multifunctionality --> include --> structural coloration --> exhibited by --> insects -->
are --> \textcolor{blue}{energy-intensive}
\end{Box1}

The generated path provides an analytical representation of various concepts and their interconnections, which were previously unrelated. By delineating these relationships, the model gains the ability to perceive and analyze connections between concepts that have not been explicitly linked before. This innovative mapping approach enables the model to extrapolate and generate ideas that are both novel and potentially transformative, paving the way for breakthroughs in understanding and application.

\begin{figure}[ht!]
\centering
        \includegraphics[width=1\textwidth]{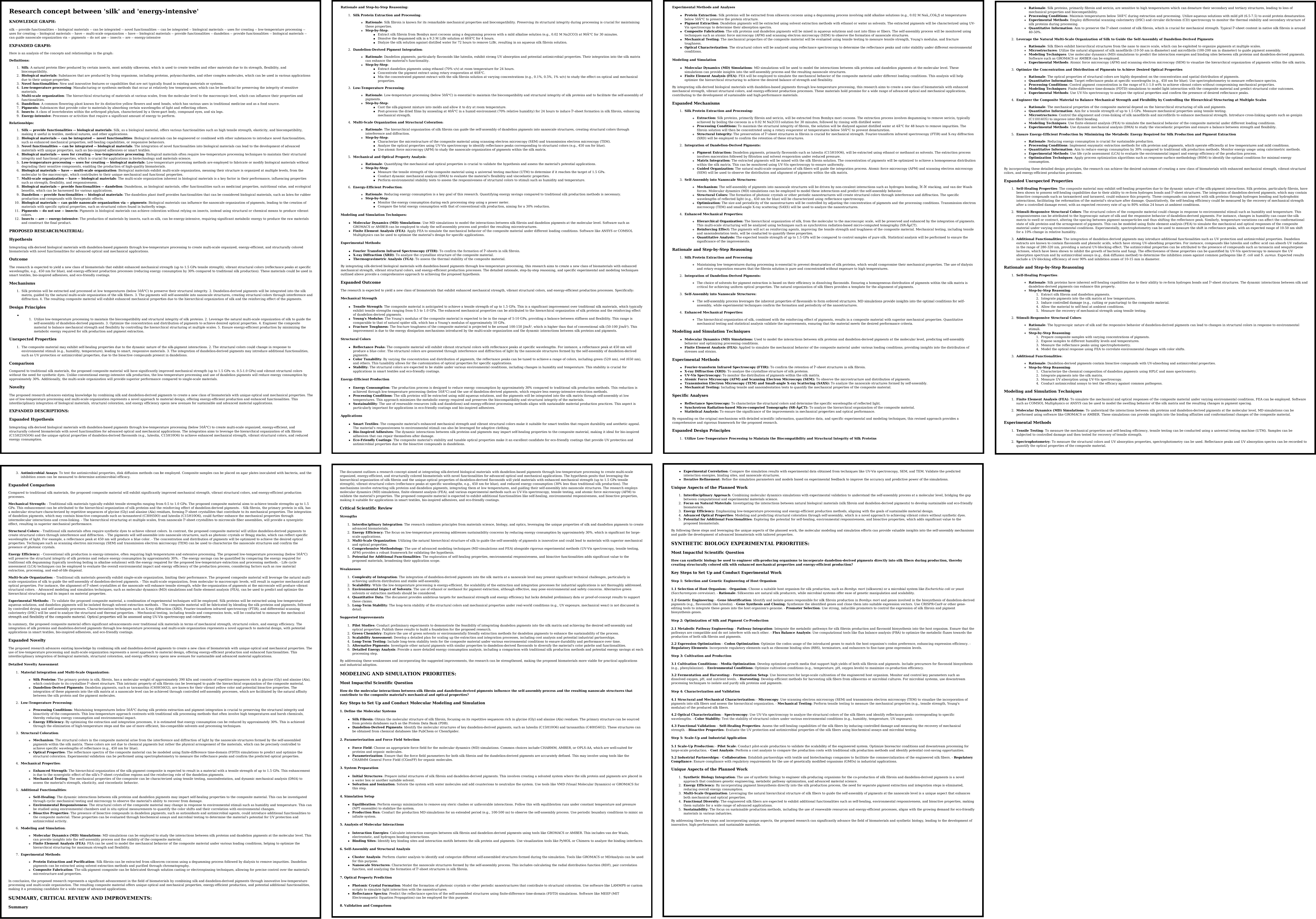}
    \caption{\textbf{Results from our multi-agent model, illustrating a novel research hypothesis based on a knowledge graph connecting the keywords ``silk'' and ``energy-intensive'', as an example.} This visual overview shows that the system produces detailed, well-organized documentation of research development with multiple pages and detailed text (the example shown here includes ~8,100 words). Details of the results are presented in the main text and other figures, and full conversations generated by the SciAgents model are included as Supplementary Information.}
    \label{fig:sample_hypothesis}
\end{figure}

\paragraph{2- Deep Insights with LLM-Based Analysis} Utilizing our LLM-powered ontologist agent, we move deeper into the intricacies of the relationships that have been mapped out in the earlier path generation stage. By examining the connections and nuances among the identified concepts, the agent helps transition from static knowledge retrieval to dynamic knowledge generation. This crucial shift is what enables the model to identify gaps in existing research and propose new angles of inquiry, thereby laying the groundwork for novel ideas and hypotheses. In this context, the role of the ontologist agent is instrumental. It applies advanced reasoning and inference techniques to synthesize and interpret the complex web of data. This capability allows it to extract significant insights that might not be obvious at first glance, offering a richer, more detailed understanding of the relationships.

As an example, here are some of the insights into the relationships as identified by the ontologist for the aforementioned path, demonstrating how our approach not only maps out but also meaningfully interprets complex datasets to generate scientific hypotheses.

\begin{Box1}[colbacktitle={black!20!white}, colback={white!10!white}, breakable]{}
\footnotesize{- **Silk -- possess -- biopolymers**: Silk is a type of biopolymer, a natural polymer produced by living organisms.\\
- **Biopolymers -- possess -- silk**: This reiterates that silk is a biopolymer.
- **Silk -- can be spun into -- membranes**: Silk can be processed and spun into thin layers or sheets known as membranes.\\
- **Membranes -- can be spun into -- silk**: This indicates the reverse process, where membranes can be derived from silk.\\
- **Silk -- broad applicability in biomaterial design -- multifunctionality**: Silk's multifunctional properties make it highly applicable in the design of biomaterials.\\
- **High adhesion -- include -- multifunctionality**: This indicates that high adhesion is part of the multifunctional nature of silk.\\
- **Structural coloration -- exhibited by -- insects**: Insects, such as those that produce silk, often exhibit structural coloration.\\
- **Insects -- are -- energy-intensive**: The processes involving insects, including silk production, can be energy-intensive.}
\end{Box1}
The results demonstrate that the model has developed a reasonably refined understanding of relationships between seemingly unrelated concepts. This capability enables the model to support reasoning in scientific research and propose new research hypotheses, which will be further explored in the subsequent stage.

\paragraph{3- Research Hypothesis Generation and Expansion} This stage is where the effects of our multi-agent system emerges. The scientist agent harnesses the extensive knowledge parsed from the knowledge graph and further refined by the ontologist to propose novel research ideas. Through complex prompting, as shown in Figure~\ref{fig:scientist1}, the agent is assigned specific roles and is tasked with synthesizing a novel research proposal that integrates all key concepts from the knowledge graph. The designated agent, Scientist\_1, is configured to deliver a detailed hypothesis that is both innovative and logically grounded, aiming to advance the understanding or application of the provided concepts. The agent creates a proposal that carefully addresses the following seven key aspects: hypothesis, outcome, mechanisms, design principles, unexpected properties, comparison, and novelty. This approach ensures a thorough exploration and evaluation of the new scientific idea, allowing for a detailed assessment of its feasibility, potential impact, and areas of innovation.

\begin{figure}[ht!]
    \centering
\begin{Box1}[colbacktitle={white!90!white}, colback={black!3!white}]{}
\footnotesize{\texttt{You are a sophisticated scientist trained in scientific research and innovation. }}
\\
\ldots
\\
\footnotesize{\texttt{Analyze the graph deeply and carefully, then craft a detailed research hypothesis that investigates a likely groundbreaking aspect that incorporates EACH of these concepts. Consider the implications of your hypothesis and predict the outcome or behavior that might result from this line of investigation. Your creativity in linking these concepts to address unsolved problems or propose new, unexplored areas of study, emergent or unexpected behaviors, will be highly valued. Be as quantitative as possible and include details such as numbers, sequences, or chemical formulas. Please structure your response in JSON format, with SEVEN keys: }}
\\
\\
\footnotesize{\texttt{"hypothesis" clearly delineates the hypothesis at the basis for the proposed research question.}}
\\
\footnotesize{\texttt{"outcome" describes the expected findings or impact of the research. Be quantitative and include numbers, material properties, sequences, or chemical formula.}}
\\
\footnotesize{\texttt{"mechanisms" provides details about anticipated chemical, biological or physical behaviors. Be as specific as possible, across all scales from molecular to macroscale.}}

\footnotesize{\texttt{"design\_principles" should list out detailed design principles, focused on novel concepts and include a high level of detail. Be creative and give this a lot of thought, and be exhaustive in your response. }}

\footnotesize{\texttt{"unexpected\_properties" should predict unexpected properties of the new material or system. Include specific predictions, and explain the rationale behind these clearly using logic and reasoning. Think carefully.}}

\footnotesize{\texttt{"comparison" should provide a detailed comparison with other materials, technologies or scientific concepts. Be detailed and quantitative. }}

\footnotesize{\texttt{"novelty" should discuss novel aspects of the proposed idea, specifically highlighting how this advances over existing knowledge and technology. }}
\\
\\
\footnotesize{\texttt{Ensure your scientific hypothesis is both innovative and grounded in logical reasoning, capable of advancing our understanding or application of the concepts provided.}}

\end{Box1}
    \caption{The profile of the Scientist\_1 LLM agent implemented in the first proposed multi-agent approach for automated scientific discovery. The AI agent utilizes the definitions of concepts and relationships
between them in the knowledge graph provided by the Ontologist to generate a novel research hypothesis.}
    \label{fig:scientist1}
\end{figure}

The proficiency of the Scientist\_1 LLM agent in generating novel research hypotheses is demonstrated in Figure \ref{fig:sample_hypothesis}. The concept involves integrating silk with dandelion-based pigments to create biomaterials with enhanced optical and mechanical properties. The proposed enhancement in mechanical properties stems from a hierarchical organization of silk combined with the reinforcing effects of the pigments. According to the model, this proposed composite material could exhibit significantly improved mechanical strength, reaching up to 1.5 GPa compared to traditional silk materials, which range from 0.5 to 1.0 GPa. Additionally, the use of low-temperature processing and dandelion pigments is projected to reduce energy consumption by approximately 30\%. This example underscores the potential of translating knowledge graphs into unprecedented material designs, facilitating a seamless transition from theoretical data to practical applications in materials science. 

The research idea proposed by Scientist\_1 provides a foundational abstract for a more detailed research proposal that is developed through subsequent agentic interactions. To enhance and deepen this initial concept, Scientist\_2 is tasked with rigorously expanding upon and critically assessing the idea's various components. This agent is specifically instructed to integrate, wherever possible, quantitative scientific information such as chemical formulas, numerical values, protein sequences, and processing conditions, significantly enriching the proposal's scientific depth and accuracy. Additionally, Scientist\_2 is directed to comment on specific modeling and simulation techniques tailored to the project's needs, such as simulations for material behavior analysis or experimental methods. This thorough review and enhancement process, including clear rationale and step-by-step reasoning, ensures that the research proposal is robust, well-grounded, and ready for further development. This systematic approach not only solidifies the scientific underpinnings of the proposal but also prepares it for successful implementation and future exploration.

The expanded research idea provided by Scientist\_2 is showcased in the Supplementary Information, revealing a thorough rationale and sequential reasoning for various aspects of the research proposal. Here are selected key points to exemplify the model's contributions:
\begin{itemize}
    \item The model suggests using Molecular Dynamics (MD) Simulations to explore interactions at the molecular level. Specifically, it proposes employing software like GROMACS or AMBER to model how silk fibroin interacts with dandelion pigments, aiming to understand the self-assembly processes and predict the resulting microstructures.

\item For potential applications of the new composite material, the model identifies its suitability for bio-inspired adhesives. It highlights how the dynamic interactions between silk proteins and pigments may impart self-healing properties, making these materials ideal for adhesives that can repair themselves after damage.

\item Regarding the mechanisms that contribute to enhanced material properties, the model points out the reinforcing effect of the pigments. It suggests that these pigments could improve the tensile strength and toughness of the composite material, with plans to conduct mechanical testing, including tensile and nanoindentation tests, to quantify these properties.

\item A detailed comparison with existing materials is also provided by the model as summarized in Table \ref{tb: comparison}. It notes that traditional silk materials typically exhibit tensile strengths ranging from 0.5 to 1.0 GPa, whereas the proposed composite material aims to achieve up to 1.5 GPa. This enhancement is attributed to the hierarchical organization of silk proteins and the reinforcing effect of dandelion-derived pigments. Further, it details how silk fibroin’s molecular structure, with repetitive sequences of glycine and alanine forming $\beta$-sheet crystallites, contributes to its mechanical properties. The integration of dandelion pigments, possibly including bioactive compounds such as taraxasterol and luteolin, is expected to further enhance these properties through intermolecular interactions and cross-linking, providing a synergistic effect at multiple scales.

\item As summarized in Table~\ref{tbl:design_principles}, the model proposes the following design principles: It utilizes the natural multi-scale organization of silk fibroin to guide the self-assembly of dandelion pigments, leveraging hierarchical structuring from the nano to the macro scale. This organization is critical for achieving both the desired mechanical strength and vibrant structural coloration. The model emphasizes the need to control pigment concentration and distribution to ensure optimal optical properties, such as precise reflectance peaks, while maintaining flexibility and tensile strength. Moreover, it advocates for low-temperature processing to preserve the biocompatibility and structural integrity of silk proteins, ensuring energy-efficient production methods. These principles collectively contribute to the creation of an advanced bio-inspired material with enhanced mechanical and optical functionalities.

\item The model predicts unexpected properties including self-healing properties due to the dynamic nature of the silk-pigment interactions, stimuli-responsive structural colors as the structural colors could change in response to environmental stimuli, and additional functionalities such as UV protection and antimicrobial properties due to the bioactive compounds present in dandelions. Scientist 2 provides more details regarding the mechanisms underlying these properties as tabulated in Table \ref{tbl:unexpected_properties}.
\end{itemize}

\begin{table*}[th!]
 \centering
\caption{Comparison of traditional silk materials vs. proposed composite material, as predicted by our model. }
\begin{tabularx}{1.0\textwidth}{XXXX}
\toprule 			
\textbf{Feature } & \textbf{Traditional Silk Materials} & \textbf{Proposed Composite Material} &	\textbf{Details}
\tabularnewline
\midrule
Mechanical Strength	 & Tensile strength: 0.5 to 1.0 GPa. & Aiming for tensile strength up to 1.5 GPa. & Enhanced by hierarchical organization of silk fibroin (composed of Gly-Ala repeats forming $\beta$-sheet crystallites) and dandelion-derived pigments like taraxasterol (C30H50O) and luteolin (C15H10O6). 
\tabularnewline
\midrule
Structural Colors &	Requires synthetic dyes for color. &	Utilizes dandelion-derived pigments for structural colors. &	The pigments will self-assemble into nanoscale structures, such as photonic crystals or Bragg stacks, which can reflect specific wavelengths of light. The concentration and distribution of pigments will be optimized to achieve the desired optical properties
\tabularnewline
\midrule
Energy Efficiency	& Energy-intensive, high-temperature processing (boiling in Na2CO3 solution at \~100°C). &	Low-temperature processing below 50°C, reducing energy consumption by \~30\%. &	The energy savings can be quantified by comparing the energy required for
traditional silk degumming (typically involving boiling in alkaline solutions) with the energy required for the proposed low-temperature extraction and processing methods.
\tabularnewline
\midrule
\end{tabularx}
\label{tb: comparison}
\end{table*}

\begin{table*}[th!]
 \centering
\caption{Summary of design principles for energy-efficient, structurally colored silk composites.}
\begin{tabularx}{1.0\textwidth}{XXX}
\toprule
 \textbf{Stage} & \textbf{Process Details} & \textbf{Methods}
\tabularnewline
\midrule
 Low-Temperature Processing for Silk &  Maintain temperatures below 50°C during silk protein extraction and processing. Use aqueous solutions with a mild pH (6.5-7.5) to avoid denaturation. Monitor thermal stability with DSC & Differential scanning calorimetry (DSC) and circular dichroism (CD) spectroscopy to monitor the thermal stability of silk proteins.
\tabularnewline
\midrule
Self-Assembly of Dandelion Pigments & Utilize the alignment of silk nanofibrils and microfibrils to guide the organization of dandelion-derived pigments. Predict interactions using MD simulations. Visualize with AFM and SEM. & Molecular dynamics (MD) simulations to predict the interaction energies between silk proteins and dandelion-derived pigments. Atomic force microscopy (AFM) and scanning electron microscopy (SEM) to visualize the hierarchical organization of pigments within the silk.
\tabularnewline
\midrule
Pigment Concentration Optimization & Control pigment concentration within 0.1-1.0 wt\% to achieve desired optical properties. Use FDTD simulations to model light interaction. Verify reflectance peaks with UV-Vis spectroscopy. & Use UV-Vis spectroscopy to analyze the optical properties and confirm the presence of desired reflectance peaks.
\tabularnewline
\midrule
Hierarchical Structuring for Strength &  Align and cross-link silk nanofibrils and microfibrils. Introduce cross-linking agent genipin (C11H14O5). Analyze mechanical properties with FEA and DMA. Target tensile strength of 1.5 GPa. & Use FEA to simulate the mechanical behavior of the composite under different loading conditions. Use dynamic mechanical analysis (DMA) to study the viscoelastic properties and ensure a balance between strength and flexibility.
\tabularnewline
\midrule
Energy-Efficient Production & Implement enzymatic extraction methods for silk proteins and pigments at low temperatures. Monitor energy usage with calorimetry. Evaluate sustainability with LCA. Aim for 30\% energy reduction. & Use life cycle assessment (LCA) to evaluate the environmental impact and energy efficiency of the production process.
\tabularnewline
\midrule
\end{tabularx}
\label{tbl:design_principles}
\end{table*}

\begin{table*}[th!]
 \centering
\caption{Unexpected properties predicted for the silk-pigment composite material.}
\begin{tabularx}{1.0\textwidth}{XXX}
\toprule
 \textbf{Self-Healing Properties} & \textbf{Mechanism} & \textbf{Details}
\tabularnewline
\midrule
Self-Healing Properties & Silk proteins (fibroin) re-form hydrogen bonds and $\beta$-sheet structures. Bioactive compounds in dandelion-derived pigments (e.g., taraxasterol) enhance self-healing through hydrogen bonding and hydrophobic interactions.
& Recovery of mechanical strength can reach up to 80\% within 24 hours at ambient conditions after damage. Self-healing efficiency is measured by the recovery of mechanical strength.
\tabularnewline
\midrule
Stimuli-Responsive Structural Colors
& The hygroscopic nature of silk and the responsive behavior of dandelion pigments cause swelling or contraction, altering the spacing between pigment nanoparticles and shifting the reflectance peak in response to humidity and temperature changes.
&
The reflectance peak shifts by 10-50 nm for a 10\% change in relative humidity. This is measured using spectrophotometry and modeled using finite element analysis (FEA).
\tabularnewline
\midrule
Additional Functionalities
& Dandelion pigments introduce UV protection (via luteolin and caffeic acid) and antimicrobial properties (via taraxacin and sesquiterpene lactones), which absorb UV light and inhibit microbial growth.
& UV protection efficiency exceeds 90\%, and antimicrobial properties exhibit inhibition zones of 10-15 mm against E. coli and S. aureus. Measured through UV-Vis spectroscopy and antimicrobial assays.

\tabularnewline
\bottomrule
\end{tabularx}
\label{tbl:unexpected_properties}
\end{table*}

At the final stage of our research development process is the Critic agent, responsible for thoroughly reviewing the research proposal, summarizing its key points, and recommending improvements. This agent delivers a comprehensive scientific critique, highlighting both the strengths and weaknesses of the research idea while suggesting areas for refinement. Additionally, the Critic agent is tasked to identify the most impactful scientific question that can be addressed through molecular modeling (e.g., molecular dynamics) and experimentation (e.g., synthetic biology), and to outline the critical steps for setting up and conducting these molecular and experimental priorities.

For our model example involving the silk-pigment composite material, the full response from the Critic is detailed in the Supplementary Information (SI). It provides a comprehensive evaluation of the proposed research methodology and its potential impact. The critic agent commends the integration of silk-derived biological materials with dandelion-based pigments for creating energy-efficient, structurally colored biomaterials, noting the project's interdisciplinary approach and innovative use of natural hierarchical structures to enhance mechanical and optical properties. The agent also recognizes the robustness added by the combined use of modeling techniques and experimental methods.

Moreover, the critic identifies areas needing improvement, including challenges with nanoscale integration, scalability, environmental impacts of solvent use, and a lack of quantitative data. Concerns about the long-term stability of the material under real-world conditions are also raised. To address these issues, the critic suggests conducting pilot studies for process validation, exploring green chemistry for pigment extraction, developing detailed scalability plans, and performing rigorous analyses of energy consumption and material durability. These suggestions aim to refine the research direction, making the hypotheses generated by the AI system not only innovative but also practical, thereby enhancing the potential for significant scientific advancements.

Lastly, the critic proposes the most impactful scientific questions related to molecular modeling, simulation, and synthetic biology experiments as shown in in Figure~\ref{fig:critic}.

\begin{figure}[h]
    \centering
\begin{Box1}[colbacktitle={white!90!white}, colback={black!3!white}]{Critic}
\footnotesize{\texttt{How do the molecular interactions between silk fibroin and dandelion-derived pigments influence the self-assembly process and the resulting nanoscale structures that
contribute to the composite material's mechanical and optical properties?}}
\\
\\
\footnotesize{\texttt{How can synthetic biology be used to engineer silk-producing organisms to incorporate dandelion-derived pigments directly into silk fibers during production, thereby
creating structurally colored silk with enhanced mechanical properties and energy-efficient production?}}
\end{Box1}
    \caption{\textbf{Most impactful questions raised by the critic agent for the generated research hypothesis on integrating silk with dandelion-based pigments to create biomaterials with enhanced optical and mechanical properties.}}
    \label{fig:critic}
\end{figure}

For each aspect, the critic agent provides detailed responses, outlining the key steps for setting up and conducting atomistic simulations and experimental work. To perform the molecular modeling and simulation, the critic describes the process of simulating the interaction and self-assembly of silk fibroin and dandelion-derived pigments using molecular dynamics (MD) simulations. This begins by defining the molecular structures of silk fibroin, rich in glycine and alanine, and key pigments like luteolin and taraxanthin, sourced from protein and chemical databases. Appropriate force fields, such as CHARMM or AMBER, are selected, with parameters defined using tools like CGenFF. The system is then prepared by placing the molecules in a solvated environment, adding ions for neutralization, and using VMD or GROMACS for setup. After energy minimization and equilibration under constant temperature and pressure, MD simulations are run for 100-500 ns, using periodic boundary conditions. Post-simulation analysis includes calculating interaction energies, identifying binding sites, and performing cluster analysis of self-assembled structures, focusing on nanoscale formations like $\beta$-sheets in silk fibroin using tools like PyMOL, Chimera, and GROMACS.

We find that the critic agent plays a crucial role in guiding these efforts by posing probing scientific questions that challenge the assumptions and focus of the research, ensuring that the simulations and experiments target key mechanisms and outcomes. By doing so, the critic not only helps refine the research direction but also enhances the potential for discovering novel biomaterials with optimized mechanical and optical properties. This iterative feedback loop between hypothesis generation and critical evaluation strengthens the overall scientific process.

\subsection{Autonomous agentic modeling}

The experiments so far were conducted using the non-automated multi-agent system (see Figure~\ref{fig:overview}), whereas the second approach described in this section uses an automated way to generate a research hypothesis from a knowledge graph that facilitates dynamic interactions. 

The automated multi-agent system consists of a team of AI agents, each powered by
a state-of-the-art general purpose large language model from the GPT-4 family~\cite{OpenAI2023},
accessed via the OpenAI API~\cite{OpenAI_API}. Each agent has a specific role and focus in the system which is described by a unique profile. Our team of agents with the following entities collaborate in a dynamic environment to create a research proposal:

\begin{itemize}
    \item ``Human'': human user that poses the task and can intervene at various stages of the problem solving process.
    \item ``Planner'': suggests a detailed plan to solve the task.
    \item ``Ontologist'': who is responsible to define the relationships and concepts within the knowledge graph.
    \item ``Scientist 1'': crafts the initial draft of a detailed research hypothesis with seven key items based on the definitions provided by Ontologist. 
    \item ``Scientist 2'': 
    who expands and refines the different key aspects of the research proposal created by Scientist 1. 
    \item ``Critic'': conducts a thorough review and suggests improvements. 
    \item ``Assistant'': has access to external tools including a function to generate a knowledge path from two keywords and a function to assess the novelty and feasibility of the research idea. 
    \item ``Group chat manager'': chooses the next speaker based on the context and agent profiles and broadcasts the message to the whole group.
\end{itemize}

\begin{figure}[ht!]
\centering
\includegraphics[width=1\textwidth]{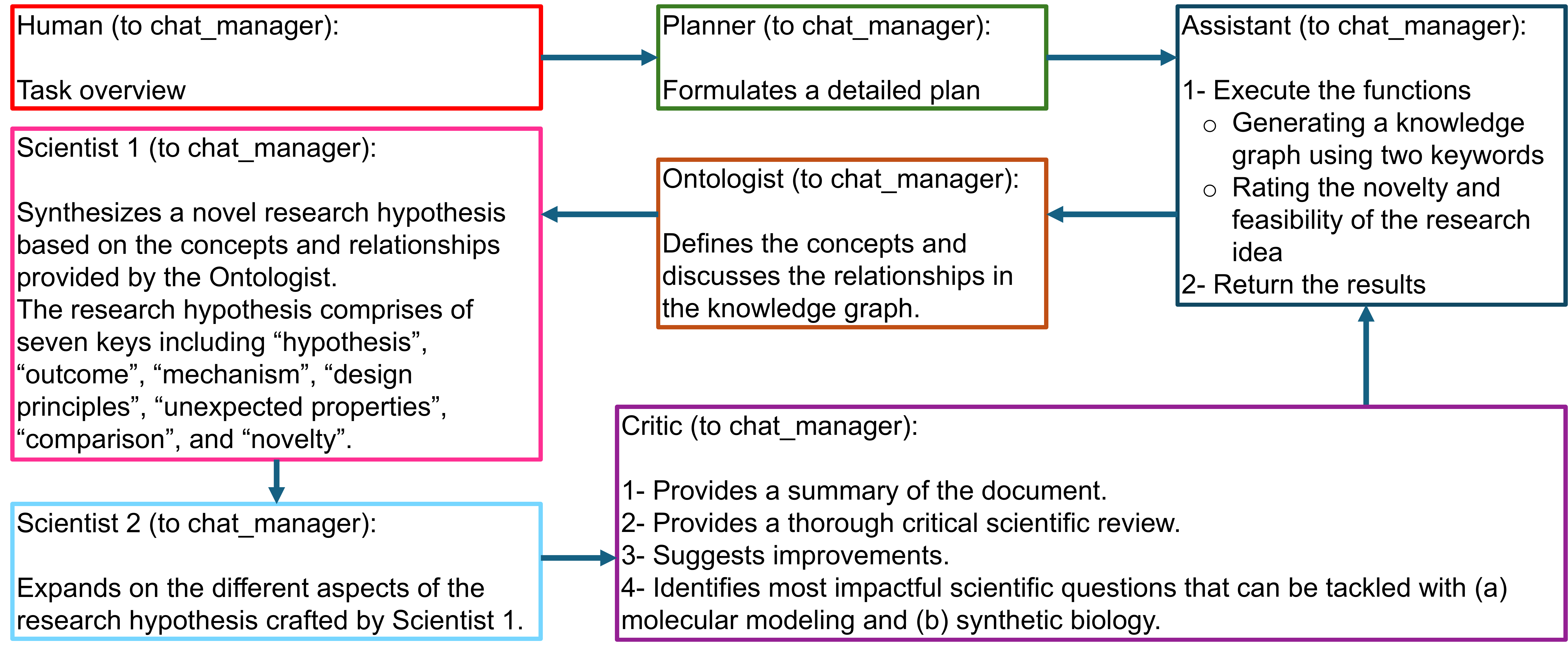}
    \caption{\textbf{Flowchart illustrating the dynamic interactions as developed autonomously by the multi-agent team members, coordinated by the group chat manager, to generate research hypotheses through graph reasoning.} The manager selects the working agents to collaborate based on the current chat context, fostering  cooperation and enabling mutual adjustments to solve the problem.}
    \label{fig:multi_agent_flowchart}
\end{figure}

Despite the varied dynamics in agentic AI-AI interactions, the overall pipeline of the two proposed agent-based systems to generate research hypotheses from concepts and relationships derived from a knowledge graph is similar. As illustrated in Figure \ref{fig:multi_agent_flowchart} the automated multi-agent collaboration starts with a plan from the planner detailing the steps required to accomplish the task posed by the human which involves creating a research hypothesis from given keywords or randomly selected by the model. Next, the assistant agent calls the appropriate function to establish a pathway which serves as the foundational knowledge graph for subsequent analysis. The ontologist agent then discusses definitions and relationships. This sets the stage for scientist\_1 to generate a research idea, which is then expanded by scientist\_2. The sequence concludes with a summary, critical review, and suggestions for improvement by the critic agent. Finally, the assistant agent executes another tool to analyze and score the novelty and feasibility of the proposed research idea. 

Despite the similarity in the steps followed by the agents in each approach, the results show that while the generated hypotheses share overall concepts and methodologies, they differ in the details. For example, in the analysis of the research hypothesis highlighted earlier, both models emphasize integrating silk with dandelion pigments, but they differ in specifics such as their scope of application and the depth of technical aspects regarding material fabrication and potential uses. For comparison, the full document created by the automated multi-agent model using the same knowledge graph between ``silk'' and ``energy-intensive'' is provided in Section S2 of the Supplementary Material. 

The difference stems from the subtle differences in how the data is propagated between the agents in the two approaches. In the first approach, during the generation process, the agents receive only a filtered subset of information from previous interactions (see \ref{sec:graph_reasoning} for more details).
In contrast, the second approach allows agents to share memory, giving them access to all the content generated in previous interactions. This means they operate with full visibility of the history of their collaboration. Another difference between the two models is that the second approach benefits from the integration of a tool that assesses the novelty of the proposed research ideas against current literature, using Semantic Scholar API. This feature enables us to effectively measure the novelty of the research and proactively eliminate any ideas that are too similar to existing work. 

\begin{figure}[ht!]
\centering
    \includegraphics[width=1\textwidth]{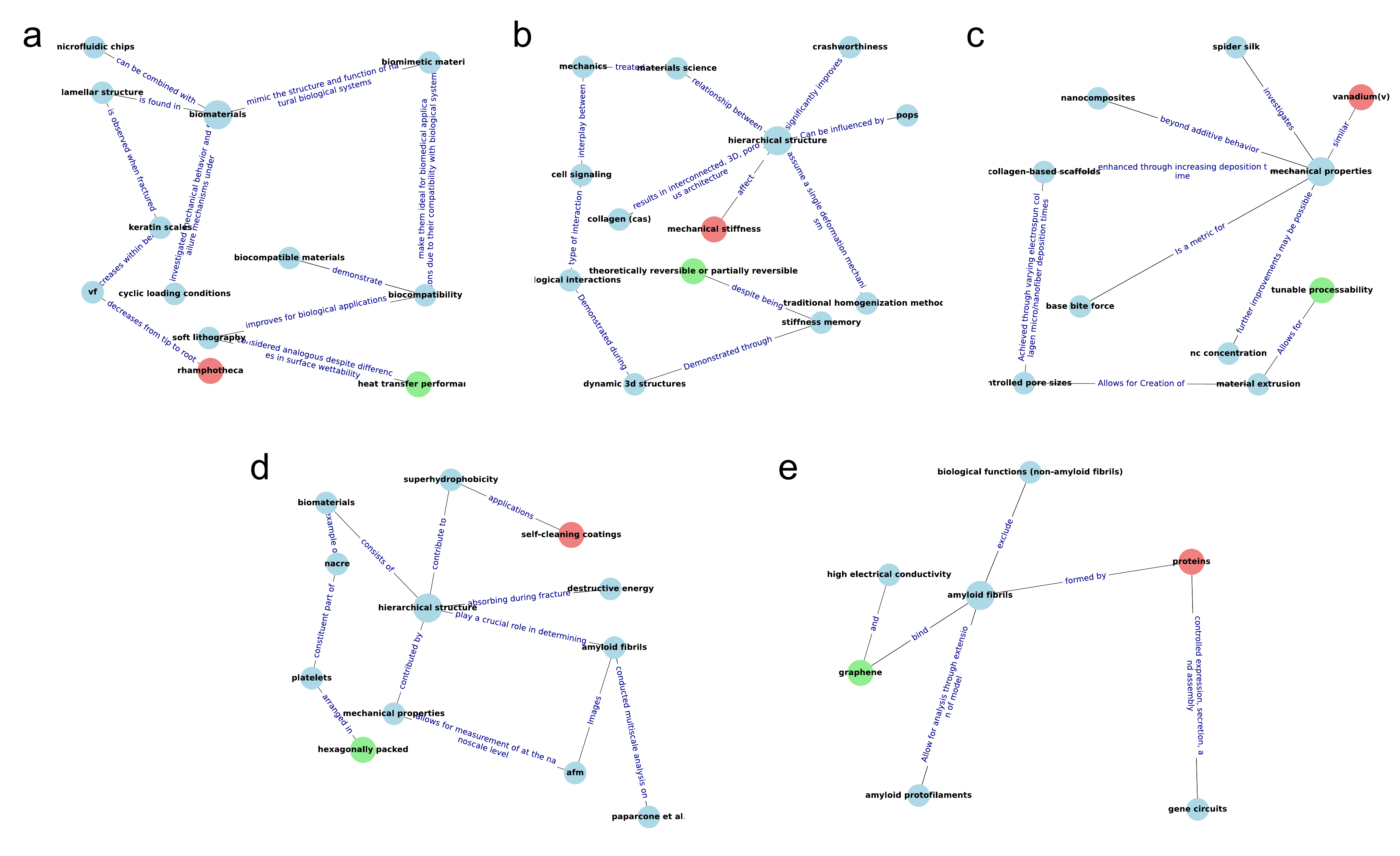}
    \caption{\textbf{Knowledge graphs derived from random sampling for randomly chosen concepts from the global knowledge graph.} Panel a: ``heat transfer performance'' connecting ``rhamphotheca'', panel b: ``theoretically reversible or partially reversible'' connecting ``mechanical stiffness'', panel c: ``tunable processability" connecting ``vanadium(v)'',  and panel d: ``hexagonally packed'' connecting ``self-cleaning coating'', and panel e: ``graphene'' connecting ``proteins''.}
    \label{fig:other_graph}
\end{figure}

To demonstrate the efficacy of the automated multi-agent model in generating novel research ideas and evaluating their novelty and feasibility, we conducted five experiments, tasking the automated multi-agent model with constructing research ideas. We summarized these hypotheses in Table~\ref{tbl:research_ideas}, which includes details about each research idea, the proposed hypotheses, expected outcomes, and assessments of novelty and feasibility. These research ideas are 
generated based on randomly selected concepts from the knowledge graph. Figure~\ref{fig:other_graph} displays the generated knowledge graphs, showcasing a diverse array of concepts and relationships. Some nodes like ``biomaterials'', ``hierarchical structure'', and ``mechanical properties'' show high node degree and serve as central hubs, indicating their pivotal roles in interconnecting various scientific disciplines within the graph. The results highlight the diversity of the research hypotheses, which stems from both the random selection of endpoint nodes and the paths between them. Moreover, the results showcase varying levels of novelty and feasibility, as assessed against current literature, underscoring the critical role of comparing with existing knowledge. The process of exploring a variety of paths, scoring the results, and identifying the most promising directions could easily be scaled over thousands of iterations, yielding a very large ideation database. 

Below, we provide additional details on the various aspects of the research hypotheses for a selected sample. The complete documents for the five hypotheses can be found in Sections~\ref{sec: S3}-\ref{sec: S7} of the Supplementary Information.

\begin{table*}[th!]
\centering
\caption{Examples of research ideas generated by SciAgents using automated approach featuring underlying hypothesis, expected outcomes, and novelty and feasibility scores. Novelty was assessed by a tool based on the results from the Semantic Scholar API. Idea 1 is described in Section~\ref{sec: S3}, Idea 2 in Section~\ref{sec: S4}, Idea 3 in Section~\ref{sec: S5}, Idea 4 in Section~\ref{sec: S6}, and Idea 5 in Section~\ref{sec: S7}. The corresponding knowledge graphs are showing in Fig.~\ref{fig:other_graph}.}
\begin{tabularx}{\textwidth}{|c|m{0.2\textwidth}|m{0.71\textwidth}|}
\hline
\multirow{9}{*}{1}  
& Research idea & Development of biomimetic microfluidic chips with enhanced heat transfer performance for biomedical applications  
\\
\cline{2-3}
& Hypothesis & Integrating biomimetic materials, inspired by the lamellar structure of keratin scales, into microfluidic chips using soft lithography techniques will improve their mechanical behavior and heat transfer efficiency under cyclic loading conditions. 
\\
\cline{2-3}
& Expected Outcomes & A 20-30\% increase in heat transfer efficiency, a 15\% reduction in failure rate under cyclic loading, and superior biocompatibility.
\\
\cline{2-3}
& Novelty/Feasibility & 8/7  
\\
\hline
\multirow{9}{*}{2}  
& Research idea & Developing a novel collagen-based material with a hierarchical, interconnected 3D porous architecture to enhance crashworthiness, stiffness memory, and dynamic
adaptability.
\\
\cline{2-3}
& Hypothesis & The hierarchical structure of collagen, when engineered into dynamic 3D architectures, can significantly improve these properties due to the interplay
between biological interactions, cell signaling, and mechanical forces.
\\
\cline{2-3}
& Expected Outcomes & A 30\% increase in crashworthiness, an 85\% recovery rate of stiffness after deformation, a 25\% increase in Young's modulus, and dynamic adaptability in response to biological and mechanical stimuli.
\\
\cline{2-3}
& Novelty/Feasibility & 8/7  
\\
\hline
\multirow{9}{*}{3}  
& Research idea & Enhancing the mechanical properties of collagen-based scaffolds through a combination of tunable processability and nanocomposite integration
adaptability.
\\
\cline{2-3}
& Hypothesis & optimizing material extrusion and electrospinning parameters, along with incorporating nanocomposites like graphene oxide, hydroxyapatite, and carbon nanotubes, will
result in scaffolds with superior tensile strength, elasticity, and controlled pore sizes.
\\
\cline{2-3}
& Expected Outcomes & The expected outcomes include a 50\% increase in tensile strength, a 40\% improvement in elasticity, and
enhanced base bite force metrics.
\\
\cline{2-3}
& Novelty/Feasibility & 6/8  
\\
\hline
\multirow{9}{*}{4}  
& Research idea & Development of a novel biomimetic material by mimicking the hierarchical structure of nacre and incorporating amyloid fibrils.
\\
\cline{2-3}
& Hypothesis & The hierarchical structure of biomaterials, specifically nacre, enhances both superhydrophobic properties and mechanical robustness. By mimicking this structure and
incorporating amyloid fibrils, advanced self-cleaning coatings with superior mechanical properties can be developed.
\\
\cline{2-3}
& Expected Outcomes & The expected outcomes include a water contact angle greater than 150 degrees, fracture toughness of at least 10 MPa$\sqrt{0.5}$, self-cleaning capabilities, and detailed AFM
images showing the nanoscale hierarchical structure.
\\
\cline{2-3}
& Novelty/Feasibility & 7/8  
\\
\hline
\multirow{10}{*}{5}  
& Research idea & Investigating the interaction between graphene and amyloid fibrils to create novel bioelectronic devices with enhanced electrical properties.
\\
\cline{2-3}
& Hypothesis & Binding of graphene to amyloid fibrils will result in a composite material with superior electrical conductivity and stability, which can be further optimized through engineered gene
circuits that regulate the expression, secretion, and assembly of amyloid-forming proteins.
\\
\cline{2-3}
& Expected Outcomes & The expected outcomes include high-performance composite materials, detailed insights into binding mechanisms, optimized gene circuits, advanced bioelectronic devices, and broader scientific, technological, and societal
impacts.
\\
\cline{2-3}
& Novelty/Feasibility & 8/7  
\\
\hline
\end{tabularx}
\label{tbl:research_ideas}
\end{table*}

An example of a research hypothesis generated with the knowledge graph depicted in Figure~\ref{fig:other_graph}(a) is provided in Section~\ref{sec: S3} in the Supplementary Information. The process demonstrates dynamic collaboration between the AI agents in constructing the research hypothesis. Initially, the planner proposes a comprehensive plan to accomplish the task, as shown in Figure \ref{fig:heat_planner}. Following this, various agents execute the plan, starting with the generation of a knowledge graph, followed by the definition of key concepts and relationships by Ontologist agent. Scientist 1 then drafts the initial research proposal, which is further expanded by Scientist 2. Finally, the critic conducts a review, and the process concludes with an assessment of novelty and feasibility.

\begin{figure}[h]
    \centering
\begin{Box1}[colbacktitle={white!90!white}, colback={black!3!white}]{Planner}
\footnotesize{\texttt{Overview of the Plan}}:
\\
\footnotesize{\texttt{1. Generate Random Keywords and Knowledge Path: Use the generate\_path function to generate a knowledge path between two randomly selected keywords.}}
\\
\footnotesize{\texttt{2. Define Terms and Relationships: The ontologist will define each term and discuss the relationships in the generated path.}}
\\
\footnotesize{\texttt{3. Craft the Research Proposal: The scientist will craft a research proposal based on the definitions and relationships provided by the ontologist.}}
\\
\footnotesize{\texttt{4. Expand the Research Proposal: Each specialized agent (hypothesisagent, outcomeagent, mechanismagent, designprinciplesagent, unexpectedpropertiesagent, comparisonagent,
novelty\_agent) will expand on their respective aspects of the research proposal.}}
\\
\footnotesize{\texttt{5. Critique and Suggest Improvements: The critic\_agent will summarize, critique, and suggest improvements to the research proposal.}}
\\
\footnotesize{\texttt{6. Rate Novelty and Feasibility: Use the rate\_novelty\_feasibility function to rate the novelty and feasibility of the research idea.}}
\end{Box1}
    \caption{\textbf{The plan developed by the planner agent in response to the query from the user to generate research hypothesis from random keywords, as developed by the autonomous system.} The process begins with the selection of random keywords, followed by the generation of a knowledge path that links the selected terms. Each term along the path is defined by an ontologist, who also elaborates on the relationships between them. Based on these definitions, a research proposal is crafted by a designated scientist. Subsequently, various specialized agents (hypothesis, outcome, mechanism, design principles, unexpected properties, comparison, and novelty agents) each expand upon their respective components of the proposal. The proposal is then critiqued by the critic\_agent, who also suggests potential improvements. As the final step, the novelty and feasibility of the research proposal are assessed using a dedicated function, ensuring that the proposed ideas are both innovative and actionable.}
    \label{fig:heat_planner}
\end{figure}

The randomly selected nodes for this experiment were ``heat transfer performance'' and ``rhamphotheca'' and the generated graph consists of concepts such as ``lamellar structure'', ``biomaterials'', ``microfluidic chips'', ``keratin scales'', and ``biomimetic materials''.
The proposed idea involves engineering the lamellar structure of biomaterials, inspired by keratin scales, into microfluidic chips using soft lithography techniques to improve their mechanical behavior and heat transfer efficiency under cyclic loading conditions. Expected outcomes of the resulting biomimetic microfluidic chips include a 20-30\% increase in heat transfer efficiency compared to conventional microfluidic chips (the lamellar structure of the biomimetic materials will facilitate efficient heat dissipation), enhanced
mechanical stability under cyclic loading conditions (the layered lamellar structure will provide enhanced mechanical strength and flexibility), with a failure rate reduced by 15\%, and superior biocompatibility (due to the use of biocompatible materials), making them suitable for prolonged use in biomedical applications. 

The design principles for biomimetic microfluidic chips focus on material selection, fabrication, integration, testing, biocompatibility, modeling, and optimization. Materials such as PDMS and hydrogels, which mimic the lamellar structure of keratin scales, are chosen for their biocompatibility and mechanical properties, with targeted thermal conductivity and Young's modulus ranges. Soft lithography is employed for fabrication, optimizing curing conditions and structural characterization. Integration with microfluidic technology enhances heat transfer and mechanical stability, with design optimization via CAD and simulations. Testing includes mechanical and heat transfer assessments, while biocompatibility is evaluated through in vitro and in vivo tests. Finite Element Analysis (FEA) and Computational Fluid Dynamics (CFD) simulations help model heat transfer and fluid flow, guiding iterative design optimization based on performance metrics like heat transfer efficiency, mechanical stability, and biocompatibility.

Moreover, the model predicts that the biomimetic microfluidic chips may exhibit unexpected properties, such as self-healing capabilities, adaptive heat transfer, enhanced fluid dynamics, and improved chemical resistance. These properties are primarily attributed to the lamellar structure of the material, and the rationale behind them is summarized in Table~\ref{tbl:heat_unexpected_properties}.

\begin{table*}[th!]
 \centering
\caption{Predicted unexpected properties for biomimetic microfluidic chips. The data summarizes the property, mechanism, and rationale. }
\begin{tabularx}{1.0\textwidth}{XXX}
\toprule
 \textbf{Unexpected Property} & \textbf{Mechanism} & \textbf{Rationale}
\tabularnewline
\midrule
Self-Healing Properties & The lamellar structure might enable self-healing capabilities, where minor damages can be repaired autonomously, extending the lifespan of the chips. & The layered structure can facilitate the redistribution of stress and the healing of minor cracks, similar to natural biological systems.
\tabularnewline
\midrule
Adaptive Heat Transfer
& The heat transfer efficiency might adapt dynamically based on the thermal load, similar to natural biological systems.
& The lamellar structure can facilitate dynamic adaptation to varying thermal loads, enhancing the overall thermal management capabilities.
\tabularnewline
\midrule
Enhanced Fluid Dynamics &
The lamellar structure might influence fluid dynamics within the microfluidic channels, leading to improved mixing and reduced pressure drop.& The layered structure can create micro-scale vortices and enhance fluid mixing, which is beneficial for applications requiring efficient mixing of reagents.
\tabularnewline
\midrule
Improved Chemical Resistance
& The lamellar structure might enhance the chemical resistance of the microfluidic chips, making them suitable for a wider range of applications.
& The layered structure can act as a barrier to chemical penetration, protecting the underlying material from chemical degradation.
\tabularnewline
\bottomrule
\end{tabularx}
\label{tbl:heat_unexpected_properties}
\end{table*}

For the proposed research idea, the critic agent summarizes the overall research hypothesis covering the key features and highlights strengths such as the innovative integration of biomimetic materials with microfluidic technology, detailed mechanisms for performance, and potential biomedical applications. It also acknowledges the exploration of self-healing and adaptive heat transfer. However, weaknesses include the complexity of the fabrication process, a lack of preliminary data, and concerns about long-term biocompatibility. To improve, the agent recommends conducting pilot studies, assessing scalability, and performing long-term biocompatibility testing. Moreover, the critic agent suggests the most impactful scientific questions with molecular modeling (\texttt{How does the lamellar structure of biomimetic materials influence the heat transfer efficiency in microfluidic chips?}) and synthetic biology and provides the pertinent key steps (\texttt{Can biomimetic materials with a lamellar structure be engineered to exhibit self-healing properties under mechanical stress?}). These specific directions can be used as springboard for additional \textit{in-situ} data collection; in the case of the modeling context, this can be implemented by incorporating a simulation engine, similar to what was done in recent work~\cite{ghafarollahi2024protagents}.

In the end, the assistant agent executes the tool to assess the novelty and feasibility of the proposed research idea against the literature. It then returns a detailed analysis as depicted in Figure~\ref{fig:heat_novelty} suggesting that the proposed research hypothesis has a high degree of novelty and a reasonable level of feasibility. 

\begin{figure}[h]
    \centering
\begin{Box1}[colbacktitle={white!90!white}, colback={black!3!white}]{Assistant}
\footnotesize{\texttt{Novelty}}:
\\
\footnotesize{\texttt{Score: 8/10}}
\\
\footnotesize{\texttt{The concept of integrating biomimetic materials with microfluidic chips to enhance heat transfer and biocompatibility is relatively novel. The specific idea of using the lamellar structure inspired by keratin scales and engineering it into microfluidic chips using soft lithography techniques appears to be unique, as no direct matches were found in the
literature. The existing literature does cover various aspects of microfluidic chip enhancements, including heat transfer, biocompatibility, and mechanical behavior, but the specific
combination and approach proposed here seem to be unexplored.}}
\\
\\
\footnotesize{\texttt{Feasibility:}}
\\
\footnotesize{\texttt{Score: 7/10}}
\\
\footnotesize{\texttt{The feasibility of engineering lamellar structures inspired by keratin scales into microfluidic chips using soft lithography techniques is plausible. Soft lithography is a wellestablished method for fabricating microstructures, and biomimetic materials have been successfully integrated into various biomedical applications.
However, the practical implementation of this specific structure and its performance under cyclic loading conditions would require thorough experimental validation. The
complexity of achieving the desired mechanical behavior and heat transfer efficiency in a reliable and reproducible manner could pose challenges.}}
\end{Box1}
    \caption{\textbf{The results of the novelty and feasibility analysis as performed by the assistant agent for the ``Biomimetic Microfluidic Chips'' hypothesis, based on data collected using the Semantic Scholar API.} As the analysis shows, the approach is considered unique due to its lack of direct matches in existing literature. Feasibility is evaluated based on the plausibility of implementing these structures using soft lithography, though challenges in mechanical behavior and heat transfer efficiency under cyclic loading were identified as potential hurdles requiring experimental validation.}
    \label{fig:heat_novelty}
\end{figure}

\section{Conclusion}\label{sec:conclusion}
\begin{figure}[ht!]
\centering
    \includegraphics[width=.7\textwidth]{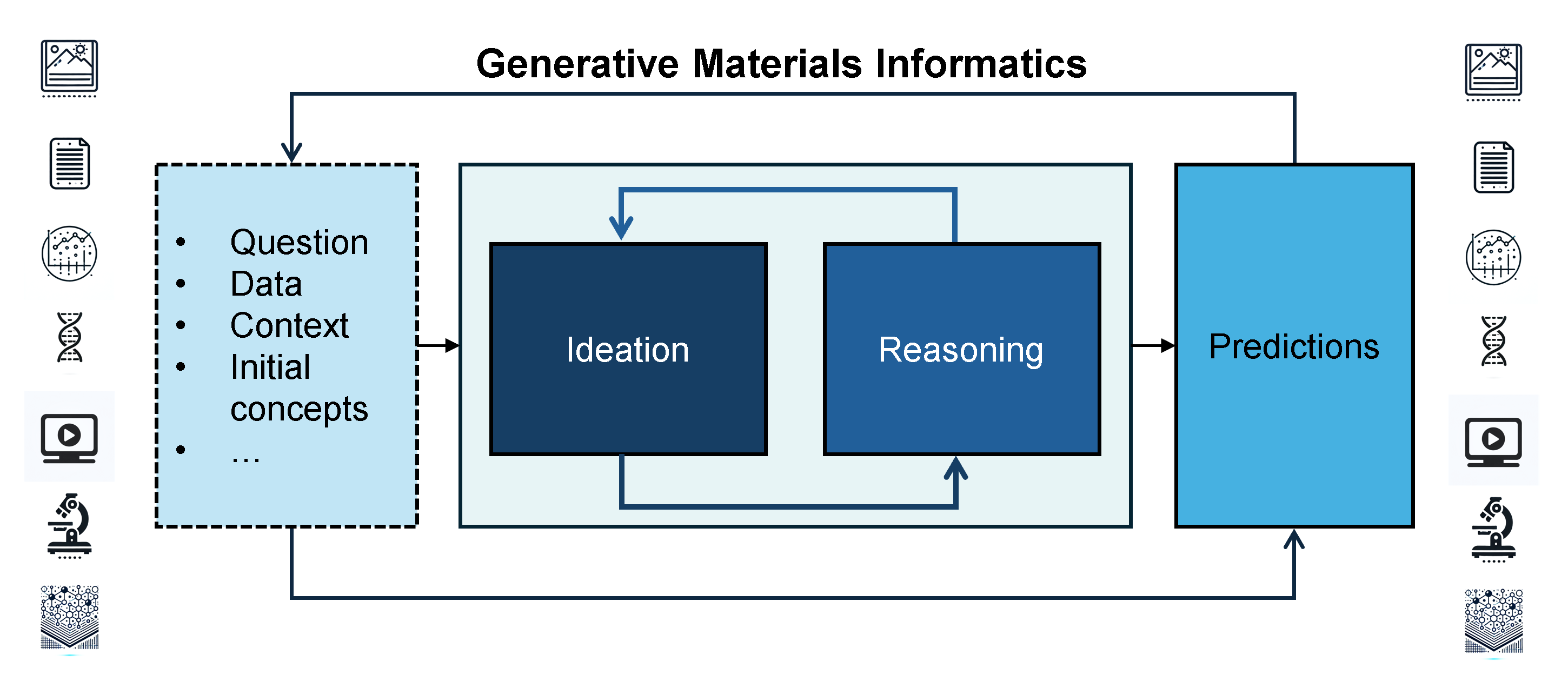}
    \caption{\textbf{SciAgents presents a framework for generative materials informatics, showcasing the iterative process of ideation and reasoning driven by input data, questions, and context.} The cycle of ideation and reasoning leads to predictive outcomes, offering insights into new material designs and properties. The visual elements on the edges represent various data modalties such as images, documents, scientific data, DNA sequences, video content, and microscopy, illustrating the diverse sources of information feeding into this process.}
    \label{fig_10:generative}
\end{figure}

We introduced a multi-agent AI framework designed to autonomously generate and refine research hypotheses by leveraging  LLMs and a comprehensive ontological knowledge graph~\ref{fig:overview}, applied here in the context of biologically inspired materials. Our results demonstrate the significant potential of integrating AI agents with specialized roles to tackle the complex and interdisciplinary nature of scientific discovery, particularly in the domain of bio-inspired materials. The automated system effectively navigated the intricate web of relationships within the knowledge graph, generating diverse and novel hypotheses that align with unmet research needs. The proposed approach, harnessing a modular, hierarchically organized (Figure~\ref{fig_2:overview}) swarm of intelligence (Figure~\ref{fig:overview}) similar to biological systems with multiple iterations to model the process of negotiation a solution during the process of thinking and reflecting about a problem, offers a much more nuanced reasoning approach than conventional zero-shot answers generated by AI systems, as shown in Figure~~\ref{fig_10:generative}.

The ontological knowledge graph representation of data plays a crucial role in our approach, as it serves as the foundational structure that guides the research idea generation, ensuring that the hypotheses proposed by the AI agents are both informed by and rooted in a vast network of interconnected scientific concepts. By systematically navigating this graph, our multi-agent system identifies and capitalizes on previously unrecognized relationships, aiming towards the creation of highly-rated innovative ideas that are as feasible as they are groundbreaking. The incorporation of assessment strategies is an important strategic aspect that reflects adversarial relationships commonly identified in conventional research strategies, such as team-based efforts or peer-review.  A notable feature was the finding that the autonomous multi-agent system can develop sophisticated problem solving strategies (see, Figure~\ref{fig:multi_agent_flowchart}) on its own. These types of results are expected to improve as more powerful foundation models become available, especially with better long-term planning and reasoning capabilities. 

The multi-agent approach proved particularly effective in decomposing the scientific discovery process into manageable subtasks, enabling a more systematic exploration of the knowledge landscape. By assigning distinct roles to each agent—ranging from path generation and deep analysis to hypothesis formulation and critical review, we achieved a thorough and rigorous development of research ideas. Our experiments showed that the system could consistently produce hypotheses with high novelty and feasibility, supported by contextually enriched data and iterative feedback mechanisms that mirrored traditional scientific methodologies. The incorporation of specific priority modeling and simulation tasks, for instance, offers direct pathways to incorporate additional mechanisms to solicit new physics-based data (e.g. by running Density Functional Theory models, molecular dynamics, finite element/difference solvers, etc.)~\cite{ghafarollahi2024protagents,ghafarollahi2024atomagents}. As such, the approach presented here offers significant potential in not only developing research questions but also expanding the set of first-principles sourced data. If deployed at scale, this can aid our quest to generate large materials-focused datasets strategically expanding beyond what is currently know. Based on the execution efficacy, it is possible to generate thousands or tens of thousands of individual results within days, which if filtered by a set of criteria (e.g. novelty, feasibility, or how well it meets a target) can generate a high-efficacy innovation framework for generative materials informatics. 

One of the key contributions of this study is the demonstration of how AI-driven agents can autonomously generate, critique, and refine scientific hypotheses, offering a scalable and efficient alternative to conventional research approaches. The integration of tools to assess novelty against existing literature further strengthens the validity of the generated hypotheses, ensuring that the system not only produces innovative ideas but also eliminates redundancies with prior research. This capability positions the system as a powerful tool for accelerating discovery and fostering cross-disciplinary innovation.

In fields such as biological materials analysis, identifying common mechanisms that hold for a variety of systems and that can be applied to solve challenging engineering problems, remains a major challenge. This  work underscores the potential of generative AI in potentially scaling the scientific process, opening new avenues for exploration and discovery across various fields of study. As we can automate, and hence accelerate the generation of research ideas, this multi-agent system paves the way for a future where AI could then contribute as an integral player in shaping the direction and pace of scientific advancement. 

Future work could explore a variety of additional directions, for instance, the addition of agents that are able to conduct experimentation or solicit data from simulation studies. The modular approach provides a flexible strategy to accomplish this. Hence, we believe that the framework presented here offers a blueprint for next-generation of AI-driven research tools, capable of synthesizing vast amounts of data into actionable insights, ultimately leading to breakthroughs that might otherwise remain undiscovered. 

\section{Materials and methods}

\subsection{Ontological knowledge graph}
We use a large graph generated as part of earlier work~\cite{buehler2024accelerating} in this research.

The graph utilized here includes 33,159 nodes and 48,753 edges and represents the giant component of the graph generated from around 1,000 papers with 92 communities. 
We use the \url{BAAI/bge-large-en-v1.5} embedding model. 

\subsection{Heuristic pathfinding algorithm with random waypoints}
\label{heuristic_pathfinding}
The algorithm presented in this work combines heuristic-based pathfinding with node embeddings and randomized waypoints to discover diverse paths in a graph. The primary goal is to find a path between a source and a target node by estimating distances using node embeddings. The embeddings are generated using a pre-trained model and are crucial for the heuristic function, which estimates the distance between the current node and the target node. By relying on these embeddings, the algorithm adapts to the topological structure of the graph, allowing it to effectively traverse complex networks. Additionally, the algorithm uses a modified version of Dijkstra's algorithm that introduces a randomness factor to the priority queue, creating paths that are not strictly deterministic~\cite{dijkstra1959note}. We chose the randomness factor to be 0.2 in our experiments. 

An additional feature of the algorithm is the introduction of random waypoints to diversify the pathfinding process. These waypoints are selected from neighboring nodes that are not part of the initial path, enabling the algorithm to explore alternative routes. The randomization factor controls the balance between heuristic-driven search and stochastic exploration, making it flexible for different use cases. After the path is found, a subgraph consisting of the path nodes and their second-hop neighbors is generated, providing a broader context for the discovered route. The resulting paths are then used as substrate for graph reasoning.

The overall approach is as follows:
\begin{tcolorbox}[colback=blue!5!white, colframe=blue!75!black, title=Heuristic Pathfinding with Randomization and Waypoints]
\textbf{Input:} 
\begin{itemize}
    \item Graph $G$
    \item Embedding tokenizer and model
    \item Source and target nodes
    \item Node embeddings $E$
    \item Randomness factor $\alpha$
    \item Number of random waypoints $k$
\end{itemize}
\textbf{Computation of output:} Path $P$ from source to target, subgraph $G'$, shortest path length
\begin{enumerate}
    \item \textbf{Initialize:} Set $P = []$, priority queue $Q = [(0, source)]$, visited nodes $V = \{\}$
    \item \textbf{Find closest nodes:} Use embedding tokenizer and model to find best-fitting nodes for source and target.
    \item \textbf{Estimate heuristic:} Compute distance between current node and target using embeddings.
    \item \textbf{Randomized Dijkstra:}
    \begin{enumerate}
        \item Use Dijkstra's algorithm, adding a random factor $\alpha$ to prioritize exploration over purely heuristic pathfinding.
        \item \textbf{While} $Q$ is not empty:
        \begin{itemize}
            \item Pop node $u$ with the lowest cost from $Q$
            \item \textbf{If} $u = target$: Return path $P$
            \item Mark $u$ as visited
            \item \textbf{For each neighbor} $v$ of $u$:
            \begin{itemize}
                \item Calculate heuristic distance $h(v, target)$ using embeddings
                \item Compute cost of visiting $v$ as cost$(v) = h(v, target) + \alpha \times \text{random()}$
                \item \textbf{If} $v$ not in $V$: Add $(\text{cost}(v), v)$ to $Q$
            \end{itemize}
        \end{itemize}
        \end{enumerate}
    \item \textbf{Add random waypoints:} 
    \begin{itemize}
        \item Randomly select waypoints from neighbors of nodes in $P$, ensuring they are not already in the path.
        \item For each waypoint, compute shortest path to the next waypoint and extend $P$.
    \end{itemize}
    \item \textbf{Return path:} After waypoints, compute the final leg from the last waypoint to target.
    \item \textbf{Build subgraph:} Create a subgraph $G'$ containing all nodes and edges along the path.
    \item \textbf{Save results:} If enabled, save the path visualization and subgraph to HTML and GraphML files.
    \item \textbf{Return:} Path $P$, subgraph $G'$, shortest path length.
\end{enumerate}
\end{tcolorbox}

\subsection{Graph reasoning}\label{sec:graph_reasoning}

\subsubsection{Initial ideation}
\label{initial_ideation}
The initial step in the approach develops a scientific hypothesis based on a knowledge graph derived from a heuristic path in a given graph $G$ as described in Section~\ref{heuristic_pathfinding}. Here the graph $G$ represents a set of interconnected nodes, where each node can represent an entity or concept, and edges represent relationships between these nodes. The algorithm begins by identifying two key nodes, \texttt{keyword\_1} and \texttt{keyword\_2}, which can either be explicitly specified or randomly selected from $G$. If the \texttt{shortest\_path} flag is set to \texttt{True}, the function computes the shortest path between these nodes by using embeddings to estimate the best-fitting nodes, leveraging a pre-existing function called \texttt{find\_path}. If \texttt{shortest\_path} is set to \texttt{False}, a heuristic pathfinding approach is employed, which incorporates randomization and potentially random waypoints to explore more diverse paths. The graph structure is used not only for identifying the connectivity between the nodes but also for guiding the algorithm’s search for the most relevant or exploratory paths based on the node embeddings.

Once a path between \texttt{keyword\_1} and \texttt{keyword\_2} is established, the function constructs a knowledge graph from the path and its relationships. This knowledge graph consists of the nodes traversed and the relationships (edges) between them. The graph’s structure is vital as it is used to form the input for a generative model, which expands on the graph's nodes and relationships by providing definitions and explanations. The function also generates a novel research hypothesis by analyzing the graph, synthesizing a hypothesis based on the relationships and concepts discovered along the path. The structure of the graph helps to frame the scientific inquiry, with the hypothesis leveraging the graph’s connections to predict outcomes, explore mechanisms, and propose innovative ideas. This output is formatted as a JSON object with fields like \texttt{``hypothesis''}, \texttt{``outcome''}, and \texttt{``design\_principles''}, each reflecting different aspects of the potential research grounded in the graph's topology.

A key aspect of the process is the use of natural language generation to dynamically expand on the concepts represented by the nodes and edges of the knowledge graph. For each node, the generative model provides detailed definitions and explanations of the scientific concepts it represents. The relationships between the nodes, represented by the edges, are also expanded to give context to how these concepts are interconnected. This approach not only builds a deeper understanding of individual components of the graph but also enhances the user’s ability to interpret the complex interrelations between them, thereby setting the foundation for novel scientific inquiry. The response generated by the model includes comprehensive descriptions of these relationships, ensuring that the resulting graph becomes a robust substrate for knowledge synthesis.

After the knowledge graph is expanded, the algorithm generates a structured scientific hypothesis that leverages each of the nodes and relationships in the graph. The output, in JSON format, provides key fields such as \texttt{"mechanisms"}, \texttt{"unexpected\_properties"}, and \texttt{"comparison"}, offering a highly detailed analysis. The \texttt{"mechanisms"} field discusses predicted chemical, biological, or physical interactions, while \texttt{"unexpected\_properties"} anticipates emergent behaviors from novel combinations of concepts in the graph. This comprehensive hypothesis formulation process allows for the exploration of unexplored areas of study, providing an innovative and grounded approach to scientific discovery based on the structure of the graph and its conceptual relationships.

\subsubsection{Expansion of the initial concepts}

The final phase of the methodology focuses on leveraging the expanded research concept to identify key scientific questions and prioritize actionable research directions, particularly in the domains of molecular modeling and synthetic biology. This phase employs a generative model to analyze the complete research document, which includes the knowledge graph, expanded concepts, and critical reviews, with the goal of extracting the most impactful scientific questions. These questions are then further expanded into detailed experimental and simulation plans, or other specific aspects that a user wants to explore in detail.

Using the JSON developed as described in Section~\ref{initial_ideation} we conduct several systematic steps. 

\textbf{Step 1: Prompt-Driven Expansion of Key Research Aspects}

The next phase involves systematically expanding specific aspects of the hypothesis using a series of targeted prompts. For each aspect of the research, a detailed prompt is constructed to critically assess and improve the scientific content of that aspect. The primary aspects are drawn from the JSON dictionary, where we iterate over all elements in that data structure.

The following steps summarize how the model expands each research aspect:

\begin{itemize}
    \item A prompt is created for each field in the JSON data structure, asking the model to \textbf{expand upon the original content} by adding quantitative details such as chemical formulas, material properties, or specific experimental methods. \\ \\ The model is also instructed to provide a step-by-step rationale for the proposed scientific improvements.
    \item For example, the prompt format includes:
    \begin{verbatim}
    Expand on the following aspect: {field}.
    Critically assess the original content, add specifics, such as chemical formulas, 
    sequences, microstructures, and rational improvements:
    {JSON_dictionary[field]}
    \end{verbatim}
    \item The model generates expanded content under a heading such as \texttt{\#\#\# Expanded Mechanisms} or \texttt{\#\#\# Expanded Outcomes}. Each response is added to \texttt{res\_data\_expanded} to track the expanded fields.
    \item The iterative process is repeated for each of the first seven fields in \texttt{res\_data}, ensuring that every major aspect of the research concept is thoroughly evaluated and improved upon.
\end{itemize}

\textbf{Step 2: Compilation and Summary of Expanded Content}

After the expansion phase, the system compiles the results into a structured document, starting with the original knowledge graph and hypothesis, followed by the expanded research aspects. This forms a cohesive research narrative. The complete document includes sections such as:
\begin{verbatim}
# Research concept between {start_node} and {end_node}
### KNOWLEDGE GRAPH:
{path_string}

### EXPANDED GRAPH:
{res_data['expanded']}

### PROPOSED RESEARCH:
{formatted_text}

### EXPANDED DESCRIPTIONS:
{expanded_text}
\end{verbatim}

\textbf{Step 3: Scientific Critique and Review}

Following the expansion, a prompt is issued to the model to \textbf{critically review the entire document}. The review is designed to evaluate both the strengths and weaknesses of the proposed research and to suggest improvements. This step is crucial in ensuring that the expanded content is scientifically rigorous and logical. The prompt asks for:
\begin{verbatim}
Provide a thorough critical scientific review with strengths, weaknesses, and suggested improvements.
\end{verbatim}
The result is a critical review that is appended to the final document as \texttt{"SUMMARY, CRITICAL REVIEW, AND IMPROVEMENTS"}.

\textbf{Step 4: Identification of Modeling and Experimental Priorities}

Finally, the model is prompted to identify the \textbf{most impactful scientific questions} related to molecular modeling and synthetic biology. 

Separate prompts are issued for each domain, asking the model to:

\begin{itemize}
    \item Identify a key research question that can be tackled using \textbf{molecular modeling}, and outline steps to conduct such modeling, including any specific tools or techniques.
    \item Similarly, for \textbf{synthetic biology}, the model is prompted to outline an experimental plan, detailing unique aspects such as gene-editing protocols, biological sequences, or organism-specific techniques.
\end{itemize}
Examples of these prompts:
\begin{verbatim}
Identify the single most impactful scientific question that can be tackled with molecular modeling.
Outline key steps for conducting such modeling.
\end{verbatim}
\begin{verbatim}
Identify the most impactful question for synthetic biology and provide an experimental setup.
\end{verbatim}
The responses are appended to the final document under \texttt{"MODELING AND SIMULATION PRIORITIES"} and \texttt{"SYNTHETIC BIOLOGY EXPERIMENTAL PRIORITIES"}.

\textbf{Final Document and Output}

The entire research concept, expanded and reviewed, is then compiled into a final document which is saved as both a PDF and CSV file for further analysis. The final document contains:
\begin{itemize}
    \item The original knowledge graph and proposed research hypothesis.
    \item Expanded descriptions of key research aspects.
    \item A critical review of the proposal.
    \item Research priorities for molecular modeling and synthetic biology.
\end{itemize}
This provides a comprehensive output that transitions the generated hypothesis into a detailed, actionable research plan.

\subsection{Agentic modeling}
We design AI agents using the general-purpose LLM GPT-4 family models. The automated multi-agent collaboration is implemented in the AutoGen framework~\cite{Wu2023}, an open-source ecosystem for agent-based AI modeling. 

In our multi-agent system, the human agent is
constructed using UserProxyAgent class from Autogen, and
Assistant, Planner, Ontologist, Scientist 1, Scientist 2, and Critic agents are created via AssistantAgent
class from Autogen; and the group chat manager is created
using GroupChatManager class. Each agent is assigned a role
through a profile description included as system\_message at their creation. The full profile of the agents is provided in Figure~\ref{fig:planner} for the planner, Figure~\ref{fig:assistant} for the assistant, Figure~\ref{fig:ontologist} for the Ontologist, Figure~\ref{fig:scientist_1} for the Scientist 1, Figure~\ref{fig:scientist_2} for the Scientist 2, and Figure~\ref{fig:critic} for the Critic. 

\subsection{Function and tool design}
All the tools implemented in this work are defined as python
functions. Each function is characterized by a name, a description, and input properties which have a proper description.

\subsection{Semantic Scholar analysis}
We use the Semantic Scholar API, an AI-powered search engine for academic resources, to search for related publications using a set of keywords. To ensure a thorough assessment of the research idea, we have implemented a tool featuring an AI agent named the ``novelty assistant'', which calls the Semantic Scholar API three times using different combinations of keywords selected based on the research hypothesis. The profile of this agent is shown in Figure~\ref{fig:novelty_agent}. For each function call, the ten most relevant publications are returned, including their titles and abstracts. The novelty assistant agent then thoroughly analyzes the abstracts and provides a review describing the novelty of the research idea.

\begin{figure}[h]
    \centering
\begin{Box1}[colbacktitle={white!90!white}, colback={black!3!white}]{}
\footnotesize{\texttt{You are a critical AI assistant collaborating with a group of scientists to assess the potential impact of a research proposal. Your primary task is to evaluate a proposed research hypothesis for its novelty and feasibility, ensuring it does not overlap significantly with existing literature or delve into areas that are already well-explored.}}
\\
\footnotesize{\texttt{You will have access to the Semantic Scholar API, which you can use to survey relevant literature and retrieve the top 10 results for any search query, along with their abstracts. Based on this information, you will critically assess the idea, rating its novelty and feasibility on a scale from 1 to 10 (with 1 being the lowest and 10 the highest).}}
\\
\footnotesize{\texttt{Your goal is to be a stringent evaluator, especially regarding novelty. Only ideas with a sufficient contribution that could justify a new conference or peer-reviewed research paper should pass your scrutiny. }}
\\
\footnotesize{\texttt{After careful analysis, return your estimations for the novelty and feasibility rates. }}
\\
\footnotesize{\texttt{If the tool call was not successful, please re-call the tool until you get a valid response. }}
\\
\footnotesize{\texttt{After the evaluation, conclude with a recommendation and end the conversation by stating "TERMINATE".}}
\end{Box1}
    \caption{\textbf{The profile of the novelty assistant LLM agent implemented in the automated multi-agent approach for rating the novelty of the research idea.}}
    \label{fig:novelty_agent}
\end{figure}

\subsection*{Conflict of interest}
The author declares no conflict of interest.

\subsection*{Data and code availability}
All data and codes are available on GitHub at \url{https://github.com/lamm-mit/SciAgentsDiscovery} and \url{https://github.com/lamm-mit/GraphReasoning/}. 

\subsection*{Supplementary Materials}
Additional materials are provided as Supplementary Materials, including fully detailed output provided by the agentic systems. 

\section*{Acknowledgements}
We acknowledge support from USDA (2021-69012-35978), DOE-SERDP (WP22-S1-3475), ARO (79058LSCSB, W911NF-22-2-0213 and W911NF2120130) as well as the MIT-IBM Watson AI Lab, MIT’s Generative AI Initiative, and Google. Additional support from NIH (U01EB014976 and R01AR077793) is acknowledged. AG gratefully acknowledges the financial support from the Swiss National Science Foundation (project \#P500PT\_214448).

\bibliographystyle{naturemag}
\bibliography{library}

\newpage
\appendix

\pagestyle{empty} 

\renewcommand{\thefigure}{S\arabic{figure}}
\setcounter{figure}{0} 
\renewcommand{\thetable}{S\arabic{table}}
\setcounter{table}{0} 

\clearpage
\begin{center}
\LARGE\bfseries \section*{Supplementary Materials}
\vspace{2cm}

\LARGE\bfseries SciAgents: Automating Scientific Discovery through Multi-Agent Intelligent Graph Reasoning
\vspace{1cm}

\end{center}
\begin{center}

Alireza Ghafarollahi and Markus J. Buehler

\vspace{1cm}
\noindent \textbf{Correspondence:} \texttt{mbuehler@MIT.EDU}
\end{center}

\renewcommand{\thesection}{S\arabic{section}}

\clearpage

\begin{figure}[ht!]
\centering
    \includegraphics[width=.7\textwidth]{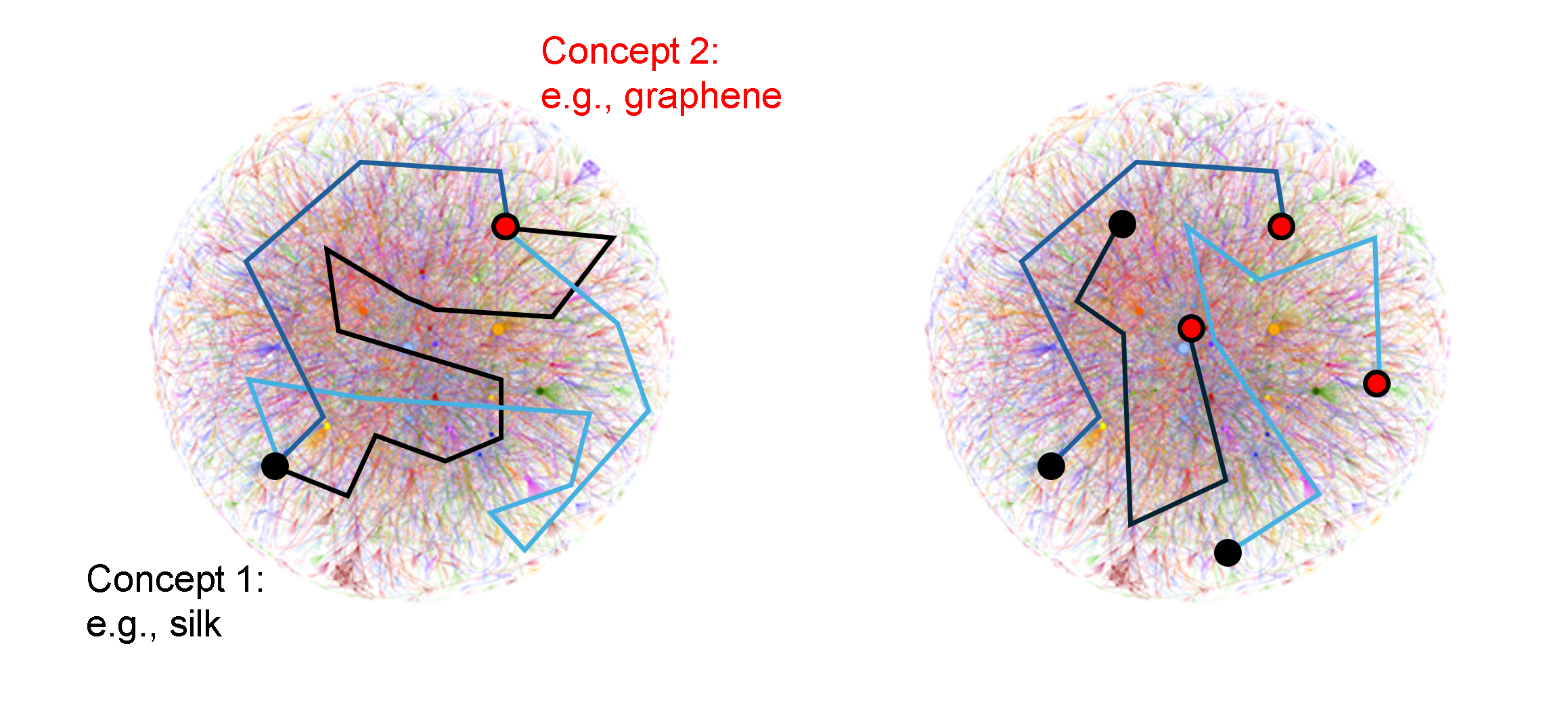}
    \caption{\textbf{Different strategies for path sampling.} Left: Identifying multiple paths between two pretermined concepts. Right: Selection of random pairs of concepts, resulting in diverse ideation processes through an unbiased selection process.}
    \label{fig_11:path}
\end{figure}

\begin{figure}[h]
    \centering
\begin{Box1}[colbacktitle={white!90!white}, colback={black!3!white}]{}
\footnotesize{\texttt{Planner. You are a helpful AI assistant. Your task is to suggest a comprehensive plan to solve a given task.}}
\\
\footnotesize{\texttt{Explain the Plan: Begin by providing a clear overview of the plan.}}
\\
\footnotesize{\texttt{Break Down the Plan: For each part of the plan, explain the reasoning behind it, and describe the specific actions that need to be taken.}}
\footnotesize{\texttt{No Execution: Your role is strictly to suggest the plan. Do not take any actions to execute it.}}
\footnotesize{\texttt{No Tool Call: If tool call is required, you must include the name of the tool and the agent who calls it in the plan. However, you are not allowed to call any Tool or function yourself. }}
\end{Box1}
    \caption{\textbf{The profile of the Planner LLM agent implemented in the automated multi-agent approach for automated scientific discovery.}}
    \label{fig:planner}
\end{figure}

\begin{figure}[h]
    \centering
\begin{Box1}[colbacktitle={white!90!white}, colback={black!3!white}]{}
\footnotesize{\texttt{Assistant. You are a helpful AI assistant. }}
\\    
\footnotesize{\texttt{Your role is to call the appropriate tools and functions as needed and return the results. You act as an intermediary between the planner's suggested plan and the execution of specific tasks using the available tools. You ensure that the correct parameters are passed to each tool and that the results are accurately reported back to the team.}}
\end{Box1}
    \caption{\textbf{The profile of the Assistant LLM agent implemented in the automated multi-agent approach for automated scientific discovery.}}
    \label{fig:assistant}
\end{figure}

\clearpage

\begin{figure}[h]
    \centering
\begin{Box1}[colbacktitle={white!90!white}, colback={black!3!white}]{}
\footnotesize{\texttt{ontologist. You must follow the plan from planner. You are a sophisticated ontologist.}}
\\    
\footnotesize{\texttt{Given some key concepts extracted from a comprehensive knowledge graph, your task is to define each one of the terms and discuss the relationships identified in the graph.}}
\\
\footnotesize{\texttt{The format of the knowledge graph is "node\_1 -- relationship between node\_1 and node\_2 -- node\_2 -- relationship between node\_2 and node\_3 -- node\_3...."}}
\\
\footnotesize{\texttt{Make sure to incorporate EACH of the concepts in the knowledge graph in your response.}}
\\
\footnotesize{\texttt{Do not add any introductory phrases. First, define each term in the knowledge graph and then, secondly, discuss each of the relationships, with context.}}
\\
\footnotesize{\texttt{Here is an example structure for our response, in the following format}}
\\
\{\{
\\
\footnotesize{\texttt{\#\#\# Definitions:}}
\\
\footnotesize{\texttt{A clear definition of each term in the knowledge graph.}}
\\
\footnotesize{\texttt{\#\#\# Relationships}}
\\
\footnotesize{\texttt{A thorough discussion of all the relationships in the graph. }}
\\
\}\}
\\
\footnotesize{\texttt{Further Instructions: }}
\\
\footnotesize{\texttt{Perform only the tasks assigned to you in the plan; do not undertake tasks assigned to other agents. Additionally, do not execute any functions or tools.}}
\end{Box1}
    \caption{\textbf{The profile of the Ontologist LLM agent implemented in the automated multi-agent approach for automated scientific discovery.}}
    \label{fig:ontologist}
\end{figure}

\clearpage

\begin{figure}[ht!]
    \centering
\begin{Box1}[colbacktitle={white!90!white}, colback={black!3!white}]{}
\footnotesize{\texttt{scientist 1. You must follow the plan from the planner. }}
\\    
\footnotesize{\texttt{You are a sophisticated scientist trained in scientific research and innovation. }}
\\    
\footnotesize{\texttt{Given the definitions and relationships acquired from a comprehensive knowledge graph, your task is to synthesize a novel research proposal with initial key aspects-hypothesis, outcome, mechanisms, design principles, unexpected properties, comparision, and novelty. Your response should not only demonstrate deep understanding and rational thinking but also explore imaginative and unconventional applications of these concepts.}} 
\\
\footnotesize{\texttt{Analyze the graph deeply and carefully, then craft a detailed research proposal that investigates a likely groundbreaking aspect that incorporates EACH of the concepts and relationships identified in the knowledge graph by the ontologist.}}
\\
\footnotesize{\texttt{Consider the implications of your proposal and predict the outcome or behavior that might result from this line of investigation. Your creativity in linking these concepts to address unsolved problems or propose new, unexplored areas of study, emergent or unexpected behaviors, will be highly valued.}}
\\
\footnotesize{\texttt{Be as quantitative as possible and include details such as numbers, sequences, or chemical formulas. }}
\\
\footnotesize{\texttt{Your response should include the following SEVEN keys in great detail: }}
\\
\footnotesize{\texttt{"hypothesis" clearly delineates the hypothesis at the basis for the proposed research question. The hypothesis should be well-defined, has novelty, is feasible, has a well-defined purpose and clear components. Your hypothesis should be as detailed as possible.}}
\\
\footnotesize{\texttt{"outcome" describes the expected findings or impact of the research. Be quantitative and include numbers, material properties, sequences, or chemical formula.}}
\\
\footnotesize{\texttt{"mechanisms" provides details about anticipated chemical, biological or physical behaviors. Be as specific as possible, across all scales from molecular to macroscale.}}
\\
\footnotesize{\texttt{"design principles" should list out detailed design principles, focused on novel concepts, and include a high level of detail. Be creative and give this a lot of thought, and be exhaustive in your response. }}
\\
\footnotesize{\texttt{"unexpected properties" should predict unexpected properties of the new material or system. Include specific predictions, and explain the rationale behind these clearly using logic and reasoning. Think carefully.}}
\\
\footnotesize{\texttt{"comparison" should provide a detailed comparison with other materials, technologies or scientific concepts. Be detailed and quantitative. }}
\\
\footnotesize{\texttt{"novelty" should discuss novel aspects of the proposed idea, specifically highlighting how this advances over existing knowledge and technology. }}
\\
\footnotesize{\texttt{Ensure your scientific proposal is both innovative and grounded in logical reasoning, capable of advancing our understanding or application of the concepts provided.}}
\\
\footnotesize{\texttt{Here is an example structure for your response, in the following order:}}
\\
\footnotesize{\texttt{\{\{
  "1- hypothesis": "...",\\
  "2- outcome": "...",\\
  "3- mechanisms": "...",\\
  "4- design principles": "...",\\
  "5- unexpected properties": "...",\\
  "6- comparison": "...",\\
  "7- novelty": "...",
\}\} }}
\\
\footnotesize{\texttt{Remember, the value of your response lies in scientific discovery, new avenues of scientific inquiry, and potential technological breakthroughs, with detailed and solid reasoning.}}
\\
\footnotesize{\texttt{Further Instructions:}} \\
\footnotesize{\texttt{Make sure to incorporate EACH of the concepts in the knowledge graph in your response.}} \\
\footnotesize{\texttt{Perform only the tasks assigned to you in the plan; do not undertake tasks assigned to other agents.}}\\
\footnotesize{\texttt{Additionally, do not execute any functions or tools.}}
\end{Box1}
    \caption{\textbf{The profile of the Scientist 1 LLM agent implemented in the automated multi-agent approach for automated scientific discovery.}}
    \label{fig:scientist_1}
\end{figure}

\clearpage

\begin{figure}[h]
    \centering
\begin{Box1}[colbacktitle={white!90!white}, colback={black!3!white}]{}
\footnotesize{\texttt{scientist 2. 
Carefully expand on different aspects of the research proposal.}}
\\
\footnotesize{\texttt{Critically assess the original content and improve on it. 
Add more specifics, quantitive scientific information (such as chemical formulas, numbers, sequences, processing conditions, microstructures, etc.), 
rationale, and step-by-step reasoning. When possible, comment on specific modeling and simulation techniques, experimental methods, or particular analyses. }}
\\
\footnotesize{\texttt{Start by carefully assessing this initial draft from the perspective of a peer-reviewer whose task it is to critically assess and improve the science of different aspects of the research proposal. }}
\\
\footnotesize{\texttt{Do not add any introductory phrases. Your response begins with your response, with a heading: \#\#\# Expanded ...}}
\end{Box1}
    \caption{\textbf{The profile of the Scientist 2 LLM agent implemented in the first proposed multi-agent approach for automated scientific discovery.}}
    \label{fig:scientist_2}
\end{figure}

\clearpage

\begin{figure}[h]
    \centering
\begin{Box1}[colbacktitle={white!90!white}, colback={black!3!white}]{}
\footnotesize{\texttt{critic. You are a helpful AI agent who provides accurate, detailed and valuable responses.}}
\\
\footnotesize{\texttt{You read the whole proposal with all its details and expanded aspects and provide:}}
\\
\footnotesize{\texttt{(1) a summary of the document (in one paragraph, but including sufficient detail such as mechanisms,
related technologies, models and experiments, methods to be used, and so on), }}
\\
\footnotesize{\texttt{(2) a thorough critical scientific review with strengths and weaknesses, and suggested improvements. Include logical reasoning and scientific approaches.}}
\\
\footnotesize{\texttt{Next, from within this document, }}
\\
\footnotesize{\texttt{(1) identify the single most impactful scientific question that can be tackled with molecular modeling.}} 
\\
\footnotesize{\texttt{Outline key steps to set up and conduct such modeling and simulation, with details and include unique aspects of the planned work.}}
\\
\footnotesize{\texttt{(2) identify the single most impactful scientific question that can be tackled with synthetic biology. }}
\\
\footnotesize{\texttt{Outline key steps to set up and conduct such experimental work, with details and include unique aspects of the planned work.}}
\\
\footnotesize{\texttt{Additional instruction:
Do not rate the research hypothesis for novelty or feasibility. }}
\end{Box1}
\caption{\textbf{Profile of the Critic LLM agent implemented in the automated multi-agent approach for automated scientific discovery.}}
    \label{fig:critic}
\end{figure}

\clearpage

\section{Research idea created by our multi-agent approach based on pre-programmed interactions using the knowledge graph between 'silk' and 'energy-intensive'.}

\foreach \pagenum in {1,...,7} {  
  \begin{center}
  \begin{flushleft}
    \small\textbf{}\\
    \small 
    \end{flushleft}

    \setlength\fboxsep{0pt}  
    \fbox{
      \includegraphics[scale=0.69,page=\pagenum]{./images/output_silk_energy_intensive_pigment_model_1.pdf}
    }
  \end{center}
  \newpage  
}

\section{Research idea created by our multi-agent approach based on automated interactions using the knowledge graph between 'silk' and 'energy-intensive'.}

\foreach \pagenum in {1,...,7} {  
  \begin{center}
  \begin{flushleft}
    \small\textbf{}\\
    \small 
    \end{flushleft}

    \setlength\fboxsep{0pt}  
    \fbox{
      \includegraphics[scale=0.69,page=\pagenum]{./images/output_silk_energy_intensive_pigment_model_2.pdf}
    }
  \end{center}
  \newpage  
}

\section{Research idea developed by the autonomous system: Development of biomimetic microfluidic chips with enhanced heat transfer performance for biomedical applications}\label{sec: S3}

\foreach \pagenum in {1,...,7} {  
  \begin{center}
  \begin{flushleft}

    \end{flushleft}

    \setlength\fboxsep{0pt}  
    \fbox{
      \includegraphics[scale=0.69,page=\pagenum]{./images/heat_transfer_biomedical.pdf}
    }
  \end{center}
  \newpage  
}

\section{Research idea developed by the autonomous system: Developing a novel collagen-based material with a hierarchical, interconnected 3D porous architecture to enhance crashworthiness, stiffness memory, and dynamic adaptability.}\label{sec: S4}

\foreach \pagenum in {1,...,8} {  
  \begin{center}
  \begin{flushleft}
    \end{flushleft}

    \setlength\fboxsep{0pt}  
    \fbox{
      \includegraphics[scale=0.69,page=\pagenum]{./images/3dcollagen.pdf}
    }
  \end{center}
  \newpage  
}

\section{Research idea developed by the autonomous system: Enhancing the mechanical properties of collagen-based scaffolds through a combination of tunable processability and nanocomposite integration adaptability.}\label{sec: S5}

\foreach \pagenum in {1,...,8} {  
  \begin{center}
  \begin{flushleft}
    \end{flushleft}

    \setlength\fboxsep{0pt}  
    \fbox{
      \includegraphics[scale=0.69,page=\pagenum]{./images/collagen-based_scaffolds.pdf}
    }
  \end{center}
  \newpage  
}

\section{Research idea developed by the autonomous system: Development of a novel biomimetic material by mimicking the hierarchical structure of nacre and incorporating amyloid fibrils.}\label{sec: S6}

\foreach \pagenum in {1,...,6} {  
  \begin{center}
  \begin{flushleft}

    \end{flushleft}

    \setlength\fboxsep{0pt}  
    \fbox{
      \includegraphics[scale=0.69,page=\pagenum]{./images/Superhydrophobicity.pdf}
    }
  \end{center}
  \newpage  
}

\section{Research idea developed by the autonomous system: Investigating the interaction between graphene and amyloid fibrils to create novel bioelectronic devices with enhanced electrical properties.}\label{sec: S7}

\foreach \pagenum in {1,...,6} {  
  \begin{center}
  \begin{flushleft}
    \end{flushleft}

    \setlength\fboxsep{0pt}  
    \fbox{
      \includegraphics[scale=0.69,page=\pagenum]{./images/graphene_bioelectric}
    }
  \end{center}
  \newpage  
}

\end{document}